\documentclass[a4paper,fleqn]{cas-dc}

\usepackage[authoryear]{natbib}
\usepackage{subcaption}

\def\tsc#1{\csdef{#1}{\textsc{\lowercase{#1}}\xspace}}
\tsc{WGM}
\tsc{QE}
\tsc{EP}
\tsc{PMS}
\tsc{BEC}
\tsc{DE}

\begin{document}

\let\WriteBookmarks\relax
\def\floatpagepagefraction{1}
\def\textpagefraction{.001}

\shorttitle{Investigating Pretext Tasks for Brain MR Segmentation}
\shortauthors{T. Nasser et~al.}

\title [mode = title]{Combining General and Domain-Specific Pretext Tasks for Brain MR Image Segmentation}

\author[1,2]{Tasneem Nasser}[type=editor]
\cormark[1]
\ead{tasneem.nasser@ucalgary.ca}

\author[2,3]{Susanne Schmid}
\ead{susanne.schmid@ucalgary.ca}

\author[2,3]{Roberto Souza}
\ead{roberto.souza2@ucalgary.ca}

\author[4]{Naser El-Sheimy}
\ead{elsheimy@ucalgary.ca}

\affiliation[1]{Biomedical Engineering, University of Calgary, Calgary, AB, Canada}
\affiliation[2]{Hotchkiss Brain Institute, University of Calgary, Calgary, AB, Canada}
\affiliation[3]{Electrical and Software Engineering, University of Calgary, Calgary, AB, Canada}
\affiliation[4]{Geomatics Engineering, University of Calgary, Calgary, AB, Canada}

\cortext[cor1]{Corresponding author}

\begin{abstract}
A key challenge in medical image analysis is the scarcity of large annotated datasets for specific populations and diseases. As deep learning models rely heavily on labeled data, effective transfer learning strategies are needed to reduce the dependence on manual annotations. Self-supervised learning has emerged as a promising approach for developing foundation models by enabling the learning of transferable feature representations from large-scale unlabeled medical imaging datasets. In this study, we investigate voxel-level brain age prediction as a domain-specific self-supervised pretext task and compare it with image inpainting, a widely used non-domain-specific alternative. We further propose a multitask self-supervised pretraining framework that jointly optimizes both objectives to learn complementary neuroimaging representations. The pretrained models are evaluated on three downstream magnetic resonance image segmentation tasks: multiple sclerosis lesion segmentation, ischemic stroke lesion segmentation, and cortical brain structure segmentation. \\
Overall, the proposed multitask pretraining framework consistently outperformed the single-task pretrained models and training from scratch across most experimental settings, demonstrating the benefit of combining domain-specific and general self-supervised learning pretext tasks for the development of generalizable neuroimaging foundation models.\\ 
\textbf{Code Availability:} The source code used in this study is publicly available at \href{https://github.com/TasneemN/Combining-General-and-Domain-Specific-Pretext-Tasks-for-Brain-MR-Image-Segmentation}{GitHub repository}.

\end{abstract}


\maketitle

\section{Introduction} \label{sec:introduction}
Data scarcity remains a major challenge in training deep learning (DL) models for medical image segmentation, where labelled data is limited due to the time-consuming nature of image annotation and the need for expert knowledge. Since most DL frameworks rely on supervised learning, they require large amounts of annotated data to learn appropriate representations.

Developing medical imaging foundation models (FMs) that learn transferable representations from diverse unlabeled medical imaging datasets has emerged as a promising solution to this challenge. Once pretrained, these models can be efficiently adapted to multiple downstream tasks using only limited annotated data \citep{he_foundation_2025}. Cross-domain generalization, which refers to the model's ability to perform well on new tasks or datasets that differ from the data it was initially trained on, for example, adapting a model trained on healthy brain scans to segment tumours or detect neurological disorders, is a key advantage of such models. A FM with this capability can be fine-tuned for downstream tasks using limited annotated data \citep{he_global-local_2021}. 
However, the success of such models relies on the ability to learn robust and transferable representations. 

Given the difficulty of manual annotation, self-supervised learning has emerged as an alternative paradigm for feature learning \citep{zheng_self-supervised_2024}. Self-supervised learning uses pretext tasks such as learning spatial relationships, predicting image rotations, and masking and then reconstructing image patches to extract useful representations from unlabeled data. Masked Autoencoder pretrains Vision Transformers by randomly masking image regions and predicting the missing content, a technique known as masked image modelling \citep{cho_domain_2025}. Kim et al. \citep{cho_domain_2025} extended this concept to Swin Transformers and demonstrated their effectiveness in learning visual representations without requiring large labelled datasets. 

Despite these advances, most existing self-supervised pretext tasks are general (\textit{i.e.}, domain-agnostic), aiming to learn general visual representations rather than anatomically meaningful neuroimaging features. Consequently, their direct application to 3D brain magnetic resonance (MR) images may not fully exploit domain-specific anatomical knowledge. Designing domain-specific pretext tasks is therefore essential to encourage models to learn representations that are more relevant to medical imaging applications \citep{cho_domain_2025}. Motivated by this, we previously proposed voxel-level brain age prediction as a domain-specific self-supervised pretext task \citep{nasser_investigating_2026}. Unlike conventional self-supervised pretext tasks that primarily exploit image statistics, voxel-level brain age prediction incorporates domain knowledge by using the assumption that the chronological age is equal to the brain age in healthy subjects  as supervision. This encourages the model to learn representations that are anatomically and biologically meaningful, capturing structural variations associated with the normal aging process. Since chronological age is routinely available as demographic metadata in normative neuroimaging datasets, it provides a practical and annotation-free source of supervision for large-scale neuroimaging pretraining.

Therefore, in this work, we aim to take a step further beyond our previous study \citep{nasser_investigating_2026}, which investigated voxel-level brain age prediction as a domain-specific pretext task for brain segmentation in data-scarce scenarios. Rather than investigating domain-specific and general pretraining tasks independently, we evaluate both single-task pretraining strategies and a multitask pretraining framework that jointly learns image inpainting and voxel-level brain age prediction.

The pretrained models are evaluated on three downstream MR image segmentation tasks. The first two tasks, multiple sclerosis (MS) lesion segmentation and ischemic stroke lesion segmentation, are not directly associated with brain aging. The last task is cortical brain structure segmentation, an age-related task. Evaluating these diverse downstream tasks allows us to investigate whether voxel-level brain age prediction learns, in addition to a general pretext task (\textit{i.e.}, image inpainting), transferable representations that generalize beyond age-related applications.

The main contribution of this work is to demonstrate that voxel-level brain age prediction, particularly when combined with multitask pretraining, serves as an effective self-supervised pretext task for brain MR image segmentation across both age-related and non-age-related downstream applications.

\section{Methodology}

\begin{figure*}[htbp]
    \centering
    \includegraphics[width=\textwidth]{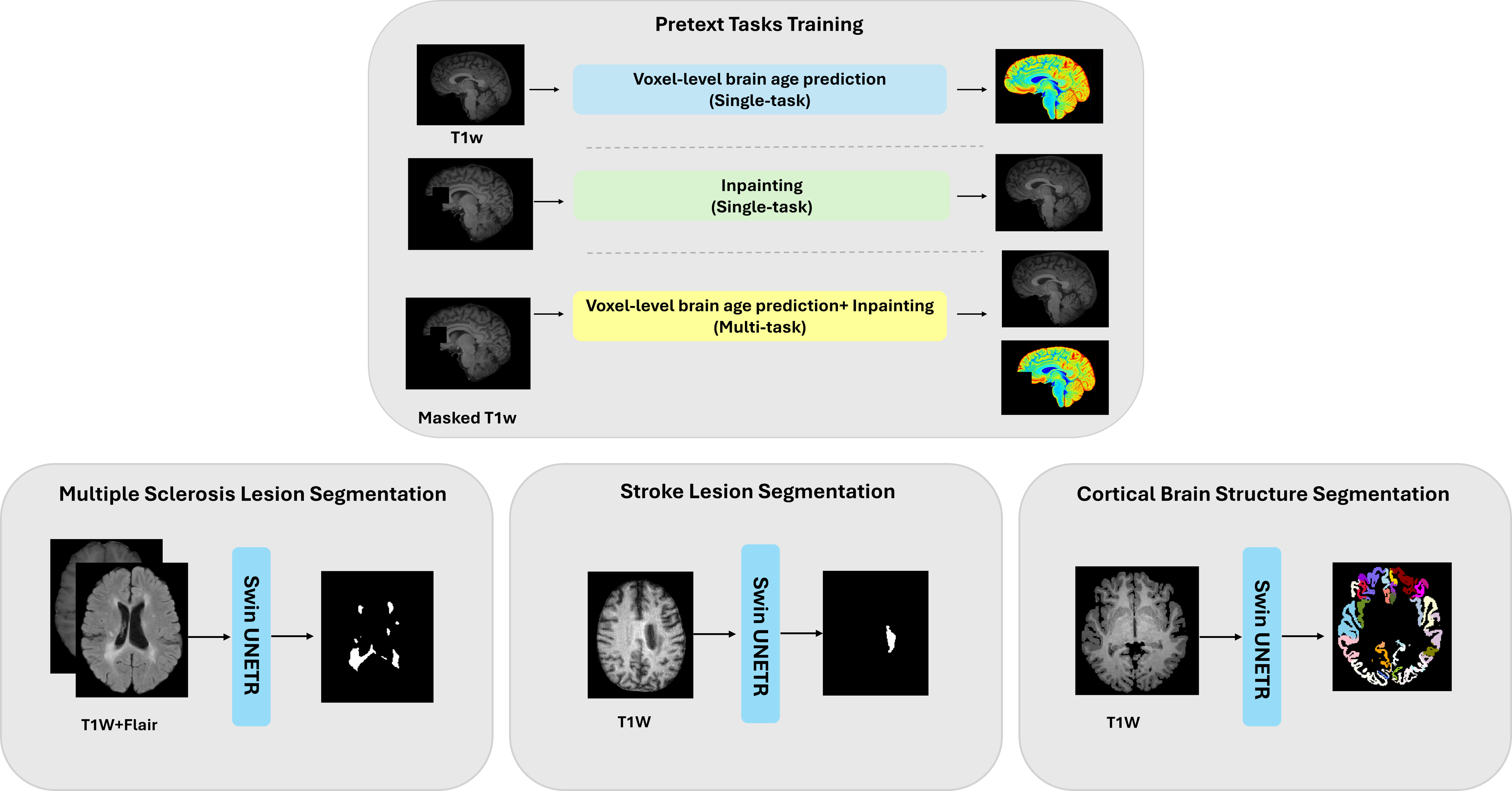} 
    \caption{Overview of the proposed framework. The figure illustrates the self-supervised pretraining strategies and their transfer to downstream MR segmentation tasks. Three pretraining strategies are investigated: (1) single-task image inpainting, in which masked MR volumes are used as input and the model reconstructs the original image; (2) single-task voxel-level brain age prediction, in which the full MR volume is used to predict a 3D voxel-wise brain age map; and (3) multitask pretraining, which jointly learns both image inpainting and voxel-level brain age prediction, producing both reconstructed MR and voxel-wise age map outputs. The pretrained SwinUNETR model is subsequently fine-tuned on three downstream segmentation tasks: multiple sclerosis lesion segmentation, and stroke lesion segmentation (non age-related), cortical brain structure segmentation (age-related)}
    \label{fig:overview}
\end{figure*}

\subsection{Proposed Framework}
This work investigates combining voxel-level brain age prediction, a domain-specific self-supervised pretext task, with image inpainting, a general pretext task, to build a brain MR foundation model. Based on preliminary architecture-selection experiments conducted using the voxel-level brain age prediction task, SwinUNETR was selected as the backbone for the proposed framework. We compare this combined approach with models pretrained using image inpainting alone and voxel-level brain age prediction alone.
We evaluate the pretrained models on three downstream MR image segmentation tasks. MS lesion segmentation and ischemic stroke lesion segmentation serve as non-age-related tasks, while cortical brain structure segmentation serves as an age-related task. This choice was deliberate, since the voxel-level brain age prediction pretext may not generalize well to non-age-related downstream tasks. Because the downstream tasks require different MR input sequences, each pretraining strategy is designed to match the sequences required by its corresponding downstream task. Figure~\ref{fig:overview} presents an overview of the proposed framework.

\begin{figure*}[htbp]
    \centering

    \begin{subfigure}{0.48\textwidth}
        \centering
        \includegraphics[width=\linewidth]{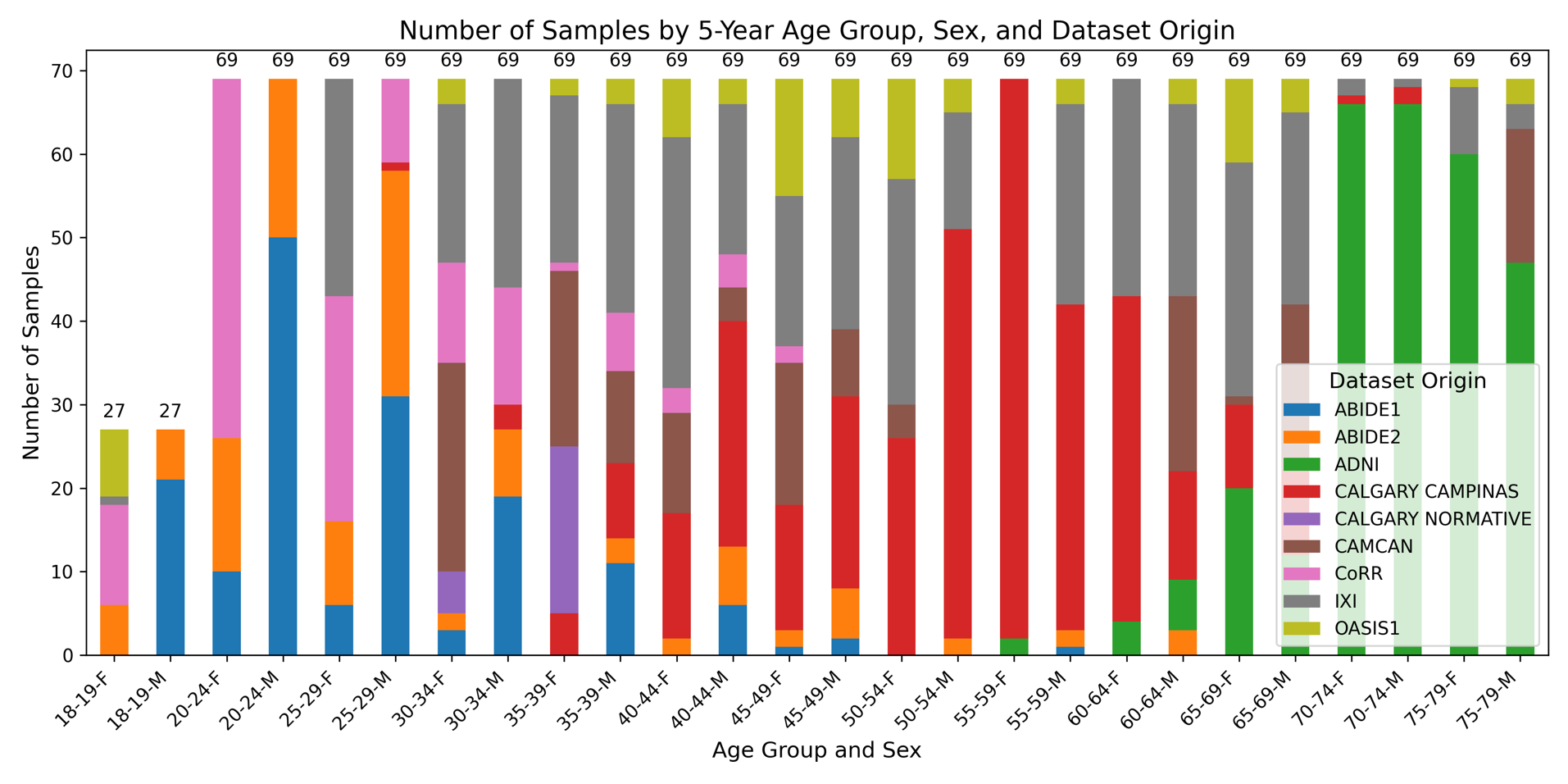}
        \caption{}
        \label{fig:multisource_distribution}
    \end{subfigure}
    \hfill
    \begin{subfigure}{0.48\textwidth}
        \centering
        \includegraphics[width=\linewidth]{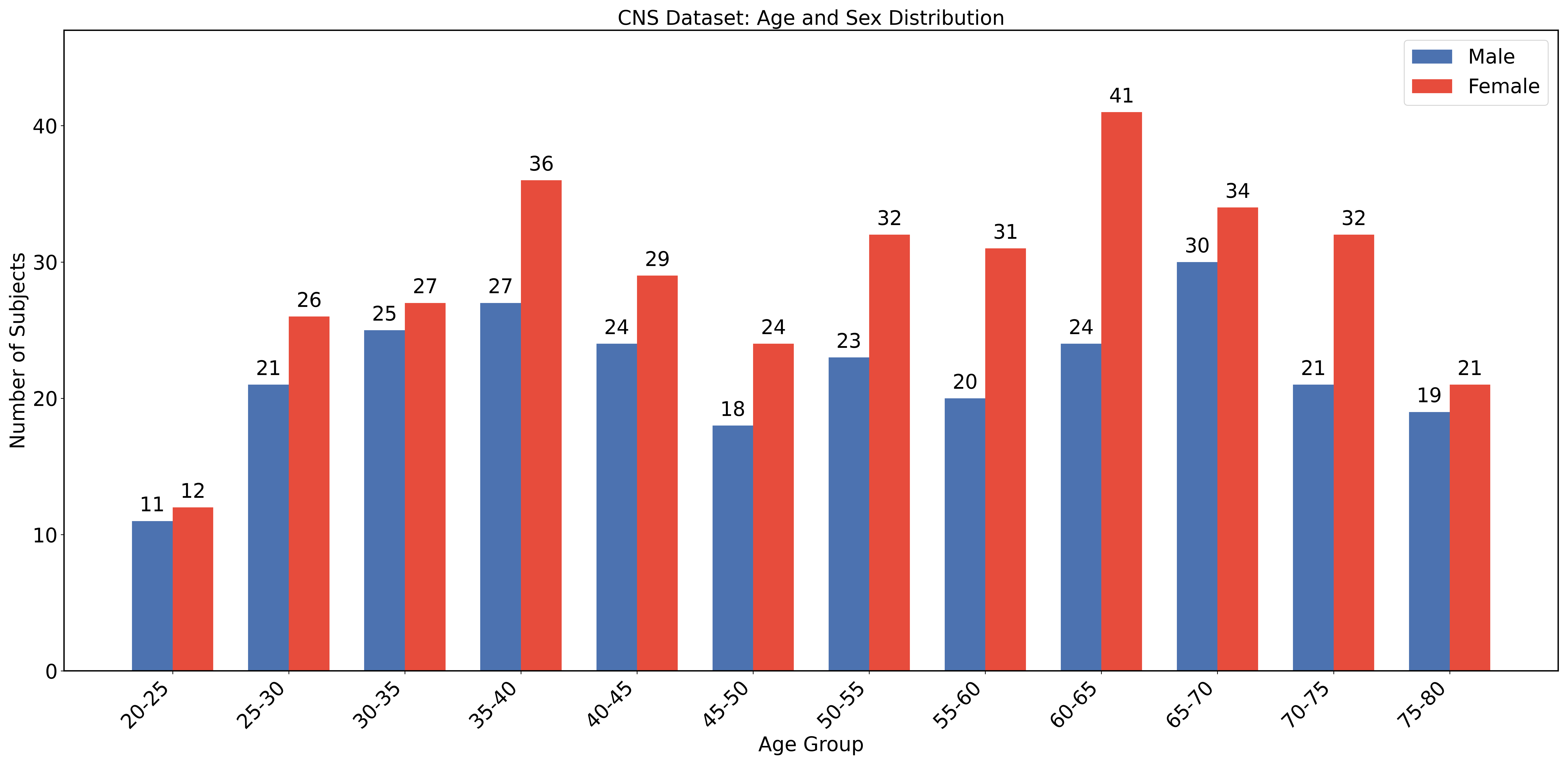}
        \caption{}
        \label{fig:cns_distribution}
    \end{subfigure}

    \caption{Age and sex distributions of the pretraining datasets. 
    (a) Multi-source T1w dataset used for pretraining prior to ischemic stroke lesion and cortical brain structure segmentation. 
    (b) Calgary Normative Study (CNS) dataset, containing paired T1w and FLAIR images, used for pretraining prior to MS lesion segmentation.}
    \label{fig:dataset}
\end{figure*}

\subsection{Datasets}


\subsubsection{Datasets for Pretext Tasks} \label{datasets pretext}

Two pretraining dataset configurations were selected according to the imaging modalities required by the corresponding downstream segmentation tasks. Paired T1-weighted (T1w) and Fluid-Attenuated Inversion Recovery (FLAIR) MR images were used for pretraining prior to MS lesion segmentation, whereas T1w-only images were used for pretraining prior to cortical brain structure and ischemic stroke lesion segmentation. A summary of the pretraining datasets, including their age and sex distributions, is presented in Figure~\ref{fig:dataset}.
Each dataset configuration was used to train the three self-supervised pretraining strategies: voxel-level brain age prediction, image inpainting, and the proposed multitask framework, before fine-tuning on the corresponding downstream segmentation tasks.

\textbf{Pretraining dataset for MS lesion segmentation}

Since MS lesion segmentation requires both T1w and FLAIR MR scans, the three self-supervised pretraining strategies were trained using paired T1w and FLAIR scans from the Calgary Normative Study (CNS) \citep{mccreary_calgary_2020} (Figure~\ref{fig:cns_distribution}). The dataset was divided into 423 training, 122 validation, and 63 test scans for self-supervised pretraining.

\textbf{Pretraining dataset for ischemic stroke lesion segmentation and brain cortical segmentation}
\\
Cortical brain structure segmentation and ischemic stroke lesion segmentation require only T1w MR scans. Therefore, we compiled a multi-source dataset of normative T1w MR scans from individuals aged 18--80 years using several publicly available datasets, as shown in Figure~\ref{fig:multisource_distribution}. These included CORR \citep{zuo_open_2014}, IXI \citep{noauthor_ixi_nodate}, ABIDE I \citep{di_martino_autism_2014}, ABIDE II \citep{di_martino_enhancing_2017}, OASIS-1 \citep{marcus_open_2007}, ADNI \citep{jack_alzheimers_2008}, Cam-CAN \citep{shafto_cambridge_2014}, the Calgary-Campinas dataset \citep{souza_open_2018}, and the CNS \citep{mccreary_calgary_2020}.

To encourage the pretrained models to learn representative features across the adult lifespan, the dataset was curated to maintain a balanced age and sex distribution across the training, validation, and test sets. This balanced sampling reduces demographic bias during pretraining and promotes the learning of more generalizable feature representations. The resulting multi-source T1w pretraining dataset comprised 1,710 MR scans, including 1,025 training, 428 validation, and 257 test scans.

\subsubsection{Dataset for MS Lesion Segmentation}

The MS lesion segmentation experiments were performed using the MSLesSeg dataset \citep{guarnera_mslesseg_2025}. The dataset comprises 115 MR scans from 75 patients with MS and includes paired T1w and FLAIR images together with expert manual lesion annotations.

\subsubsection{Dataset for Ischemic Stroke Lesion Segmentation}
We utilized the ISLES 2026 Challenge training dataset \citep{noauthor_ischemic_nodate, absher_stroke_2024}, which comprises 1,453 cases collected from multiple centers. To focus the analysis on a more stable stage of stroke pathology, cases acquired during the acute and subacute phases were excluded. During these earlier stages, stroke lesions undergo dynamic biological and structural changes, including evolving edema, inflammatory responses, and tissue remodeling, resulting in substantial variability in lesion appearance and boundaries \citep{yuh_mr_1991, cotinat_dynamics_2025, hernandez_petzsche_isles_2022}. Restricting the dataset to chronic-stage cases, where these transient processes have largely resolved, provided a more consistent setting for evaluating the impact of different pretraining strategies on downstream stroke lesion segmentation. After these exclusions, 649 chronic-stage scans remained for the experiments. 

\subsubsection{Dataset for Cortical Brain Structure Segmentation} \label{cortical dataset}
The Mindboggle-101 dataset \citep{klein_101_2012} was used for the healthy brain cortical segmentation task. The dataset consists of 101 T1w MR images from healthy participants collected from multiple publicly available neuroimaging datasets. The constituent datasets include the Nathan Kline Institute Rockland Sample (NKI-RS-22), Nathan Kline Institute Test--Retest dataset (NKI-TRT-20), Multi-Modal MRI Reproducibility Resource (MMRR-21), MMRR 3T/7T dataset (MMRR-3T7T-2), Human Language Network dataset (HLN-12), OASIS Test--Retest dataset (OASIS-TRT-20), Colin27, and additional subjects included in the original Mindboggle-101 collection.

The T1w MR images were processed using the FreeSurfer \texttt{recon-all} pipeline for cortical surface reconstruction and segmentation. Cortical regions were manually edited according to the Desikan--Killiany--Tourville (DKT) cortical labeling protocol, which was developed to provide anatomically consistent cortical parcellations based on reliable anatomical landmarks. The resulting manually labeled data provide reference cortical parcellations suitable for training and evaluating automated brain segmentation methods.

\subsection{Preprocessing}

For the CNS dataset used in the paired T1w--FLAIR pretraining configuration, an additional within-subject multimodal registration step was performed prior to model training. For each subject, the FLAIR image was rigidly registered to the corresponding T1w image using FreeSurfer's \texttt{eA\_robust\_register}, with the T1w image serving as the reference. The estimated rigid transformation was then applied to resample the FLAIR image into T1w space, ensuring spatial correspondence between the two modalities \citep{reuter_highly_2010}.
\\
For all datasets, MR scans were resampled to an isotropic resolution of 1 mm³ and their intensities were normalized to the range [0,1]. Due to GPU memory limitations, all models were trained using randomly cropped image patches of size 128×128×128.

\subsection{Model Architecture} 
Three network architectures, U-Net, UNETR, and SwinUNETR, were initially evaluated to identify the most suitable backbone for the proposed framework \citep{ronneberger_u-net_2015, hatamizadeh_unetr_2022, tang_self-supervised_2022}. U-Net relies on convolutional operations, whereas UNETR employs a Transformer-based encoder to capture global contextual information. SwinUNETR combines hierarchical feature extraction with window-based self-attention, enabling efficient modeling of both local anatomical details and long-range spatial dependencies. These characteristics make it particularly well suited for learning volumetric brain MR representations. Based on a preliminary comparison using the voxel-level brain age prediction task (Table~\ref{tab:architecture}), SwinUNETR achieved the best performance across both pretraining dataset configurations and was therefore selected as the backbone for all subsequent pretraining and downstream segmentation experiments.
\subsection{Self-Supervised Pretraining Strategies}

To develop the proposed neuroimaging FM, SwinUNETR was pretrained using three self-supervised learning strategies: image inpainting (Section \ref{Image Inpainting}), voxel-level brain age prediction (Section \ref{voxel-level brain age prediction}), and a multitask framework (Section \ref{multitask framework}) that jointly optimizes both objectives. Each strategy was independently pretrained using the datasets illustrated in Figure~\ref{fig:dataset} before being fine-tuned on the downstream segmentation tasks. The pretraining input modalities were selected to match the requirements of the corresponding downstream task. Specifically, T1w MR was used for cortical brain structure segmentation and ischemic stroke lesion segmentation, whereas concatenated T1w and FLAIR images were used for MS lesion segmentation to maintain consistency between the pretraining and downstream input modalities.
For each pretraining strategy, multiple optimization hyperparameter configurations were empirically evaluated during preliminary experiments. The best-performing configuration, selected based on the validation set, was adopted and is reported in the corresponding subsection.

\subsubsection{Image Inpainting} \label{Image Inpainting}

Image inpainting was selected as a non-domain specific self supervised pretext task because it is well established in the computer vision domain \citep{pathak_context_2016}. The objective is to reconstruct missing regions of an input MR volume by exploiting the anatomical context of the surrounding voxels, thereby encouraging the model to learn meaningful structural representations that can be transferred to downstream tasks.

Following previous work \citep{elharrouss_image_2020, cho_domain_2025, tang_self-supervised_2022}, missing regions were simulated using a coarse dropout strategy. Specifically, 12 cubic regions of size ($32 \times 32 \times 32$) were randomly removed/cropped from each input patch of size ($128 \times 128 \times 128$), masking approximately $19\%$  of the input volume. Randomly positioned masks avoid the trivial solutions associated with fixed masking patterns while exposing the model to diverse anatomical contexts during training.

The network was trained to reconstruct the original MR volume from the corrupted input. To improve reconstruction quality beyond voxel-wise similarity, a perceptual loss based on AlexNet feature representations was incorporated, encouraging the reconstructed images to preserve high-level anatomical structures in addition to local image intensities \citep{zhang_unreasonable_2018, stoean_using_2022}.

The model was trained using the Adam optimizer with an initial learning rate of $1\times10^{-3}$ and a weight decay of $1\times10^{-4}$. A StepLR scheduler was employed with a step size of 70 epochs and a decay factor of $\gamma=0.8$. The network was optimized by minimizing the perceptual reconstruction loss.

\subsubsection{Voxel-Level Brain Age Prediction} \label{voxel-level brain age prediction}

Voxel-level brain age prediction is proposed as a domain-specific self-supervised pretext task in which the network learns to predict a 3D brain age map from the input MR volume. Unlike conventional brain age prediction, which estimates a single chronological age for the entire brain, the proposed formulation assigns an age value to every voxel within the brain mask, encouraging the network to learn spatially localized age-related anatomical representations.

To generate the training labels, each voxel within the brain mask was assigned the subject's chronological age. During training, integer noise uniformly sampled from ([-2,2]) years was added to the age labels to introduce voxel-wise label variability and prevent the network from converging to the trivial solution of assigning the same age value to all voxels \citep{gianchandani_multitask_2024}. The noise was applied only during training and was removed during validation and testing, where the original chronological age was used to provide an unbiased evaluation of the learned representations. The relatively small noise range was selected to introduce sufficient variability while avoiding systematic bias in the age prediction task \citep{gianchandani_multitask_2024}.

The network was optimized using the Mean Absolute Error (MAE) loss, computed between the predicted and ground-truth voxel-wise age maps within the brain mask. By minimizing the voxel-wise MAE, the model learns age-sensitive representations that capture both local anatomical characteristics and global structural information relevant to brain aging.

The model was trained using the Adam optimizer with an initial learning rate of $1\times10^{-3}$, a weight decay of $1\times10^{-4}$, and a StepLR scheduler with a step size of 150 epochs and a decay factor ($\gamma$) of 0.6.

\subsubsection{Multitask Framework} \label{multitask framework}

Building upon our previous work \citep{nasser_investigating_2026}, we further propose a multitask self-supervised pretraining framework that jointly optimizes voxel-level brain age prediction and image inpainting. The motivation is to learn richer and more transferable feature representations by combining a domain-specific pretext task, voxel-level brain age prediction, which encourages the network to learn anatomically meaningful contextual and age-related representations, with a non-domain-specific pretext task, image inpainting, which learns contextual and structural representations by reconstructing missing image regions. By jointly optimizing both objectives, the shared SwinUNETR encoder is encouraged to learn complementary representations that capture both domain-specific anatomical characteristics and general contextual image features.

During training, the input MR volume is first masked using the same coarse dropout strategy employed for image inpainting. The same corrupted input is then provided to a shared SwinUNETR architecture, where both pretext tasks share the same encoder and decoder and differ only in their task-specific prediction heads. The reconstruction head predicts the original MR volume, while the brain age prediction head estimates the voxel-wise brain age map only for the visible (non-corrupted) brain regions. This design enables the shared network to simultaneously learn contextual structural representations through image reconstruction and contextual domain-specific anatomical representations, while preventing the brain age prediction task from relying on artificially masked regions. 

The multitask model is optimized using a weighted combination of the reconstruction and brain age prediction objectives:

\begin{equation}
L_{\text{total}}=\lambda_{\text{inp}}L_{\text{perc}}+\lambda_{\text{age}}L_{\text{MAE}}
\end{equation}

\noindent where $(L_{\text{perc}})$ denotes the perceptual reconstruction loss and $(L_{\text{MAE}})$ is the voxel-wise MAE loss for brain age prediction. The weighting coefficients were empirically set to $(\lambda_{\text{inp}}=100)$ and $(\lambda_{\text{age}}=1)$.

The multitask model was trained using the Adam optimizer with an initial learning rate of $1\times10^{-3}$ and a weight decay of $1\times10^{-4}$. The learning rate was updated using a StepLR scheduler with a step size of 150 epochs and a decay factor of $\gamma=0.8$.

\subsection{Fine-tuning for the Downstream Segmentation Tasks}

The pretrained models are evaluated on three downstream MR segmentation tasks. MS lesion segmentation and stroke lesion segmentation  are considered non-age-related tasks, whereas cortical brain structure segmentation 
is selected as an age-related task. 

For each downstream segmentation task, multiple fine-tuning hyperparameter configurations were empirically evaluated during preliminary experiments. The best-performing configuration, selected based on the validation set, was adopted and is reported in its corresponding subsection.

To evaluate the effect of pretraining under different levels of downstream data availability, each task was further assessed using multiple training-set sizes. The selected subset sizes were constrained by the total number of available subjects in each dataset and were chosen to span progressively larger training regimes, beginning with the smallest feasible subset. This design enables evaluation of the relative benefit of pretraining compared with training from scratch in low-data settings and how this benefit changes as the amount of labeled downstream data increases. In particular, it allows us to examine whether the performance advantage of pretrained models decreases, disappears, or is potentially reversed as sufficient labeled data become available for effective training from scratch.

\subsubsection{MS Lesion Segmentation} \label{MS Lesion Segmentation}

Unlike the other downstream tasks, MS lesion segmentation requires both T1w and FLAIR MR, as MS lesions are often difficult to identify on T1w images alone and are better visualized on FLAIR. Consequently, pretraining for this task was performed using the paired T1w and FLAIR images from the CNS dataset.

The downstream segmentation experiments were conducted using progressively increasing training set sizes of 11, 22, 30, 42, 50, and 58 subjects, while maintaining fixed validation and test sets of 20 and 15 subjects, respectively. For each training set size, four initialization strategies were evaluated: training from scratch, image inpainting pretraining, voxel-level brain age prediction pretraining, and the proposed multitask pretraining framework. All methods were fine-tuned using identical training, validation, and test splits to ensure a fair comparison.

Different fine-tuning strategies were investigated for the MS lesion segmentation task. First, the entire pretrained network was fine-tuned using different learning rates ($10^{-3}$, $10^{-4}$, and $10^{-5}$). We then evaluated a strategy in which the segmentation decoder was reinitialized prior to end-to-end fine-tuning of the entire network. The decoder was reinitialized to allow it to learn task-specific segmentation representations while retaining the transferable representations learned by the pretrained encoder. Additional experiments included freezing the pretrained encoder while optimizing only the reinitialized decoder and segmentation head, as well as a progressive fine-tuning strategy in which the decoder and segmentation head were first optimized before unfreezing the entire network for joint optimization. The model achieving the highest validation performance was selected, and the corresponding test results are reported.

The best-performing fine-tuning strategy, in which the segmentation decoder was reinitialized prior to end-to-end fine-tuning of the entire network, was trained using the Adam optimizer with an initial learning rate of $1\times10^{-3}$, a weight decay of $1\times10^{-4}$, and a StepLR scheduler with a step size of 120 epochs and a decay factor ($\gamma$) of 0.6. Models initialized from the self-supervised pretrained weights were fine-tuned for 300 epochs, whereas the model trained from scratch was trained for 400 epochs to allow sufficient convergence without the benefit of pretrained initialization. All other training settings were kept consistent across the evaluated strategies.

\subsubsection{Ischemic Stroke Lesion Segmentation} \label{Stroke Lesion Segmentation}
Stroke lesion segmentation was selected as an additional downstream task unrelated to brain aging, providing a more independent assessment of the transferability of the learned representations. In the present study, this task was formulated using the T1w scans available in the ISLES 2026 dataset. The T1w-only input configuration enabled all three pretraining strategies to be trained using the larger multi-source normative pretraining dataset combined from several publicly available cohorts. This provided a substantially more diverse pretraining setting than the MS lesion segmentation experiments, for which the requirement for paired T1w and FLAIR images restricted pretraining to the CNS dataset. Consequently, the stroke lesion segmentation experiments provided a complementary assessment of whether representations learned from a larger and more heterogeneous pretraining cohort could generalize to a non-age-related downstream segmentation task.

A fixed validation set of 50 scans and a fixed test set of 100 scans were used throughout all experiments to ensure direct comparability between the evaluated methods. From the remaining scans, nested training subsets containing 10, 20, 50, 100, 150, 200, 250, 300, 350, and 400 cases were constructed such that each smaller subset was fully contained within the subsequent larger subset. To ensure a fair comparison, identical training subsets were used across all training strategies at each dataset size. This experimental design simulated realistic scenarios with limited annotated data, where pretraining is expected to provide the greatest benefit, while ensuring that performance differences were attributable solely to the learning strategy rather than differences in the sampled training cohorts. The progressively increasing training subsets enabled a systematic evaluation of the effectiveness of pretraining under varying levels of labeled data availability, with particular emphasis on the low-data regime and on how the benefits of pretraining evolve as additional annotated training data become available.

The fine-tuning strategies evaluated for ischemic stroke lesion segmentation were selected based on the findings obtained from the MS lesion segmentation experiments. Specifically, all pretrained models, including voxel-level brain age prediction, image inpainting, and the proposed multitask framework, were fine-tuned by updating all network parameters. In addition, an alternative fine-tuning strategy was evaluated for the multitask framework, in which the decoder was randomly reinitialized while the encoder was initialized with the pretrained weights before fine-tuning the entire network. This strategy was included because it demonstrated best performance in the MS lesion segmentation experiments. Models trained from scratch were optimized using the same network architecture and training settings to ensure a fair comparison with their pretrained counterparts.

All models were trained for 300 epochs using the Adam optimizer with an initial learning rate of $1\times10^{-3}$, a weight decay of $1\times10^{-4}$, and a StepLR learning-rate scheduler with a step size of 120 epochs and a decay factor ($\gamma$) of 0.6. An effective batch size of 8 was used for all experiments. The model achieving the lowest validation loss was selected for evaluation on the independent test set. Each experimental configuration, defined by the combination of training-set size and training strategy, was repeated five times using different random seeds to account for stochastic variation during training. All reported quantitative results are presented as the mean ± standard deviation across the five independent runs.

\subsubsection{Cortical Brain Structure Segmentation} \label{cortical Brain Structure Segmentation}

Cortical brain structure segmentation was selected as an age-related downstream task to evaluate whether representations learned through voxel-level brain age prediction transfer effectively to the segmentation of anatomical structures that undergo systematic changes across the adult lifespan. In contrast to the MS and ischemic stroke lesion segmentation tasks, which were selected as downstream applications not directly related to brain aging, cortical segmentation provided a complementary setting for investigating the transferability of the learned representations to healthy brain anatomy. The task was formulated using the T1w scans and corresponding DKT cortical parcellations from the Mindboggle-101 dataset. As this task requires only T1w images, all three pretraining strategies were initialized using the larger multi-source normative T1w pretraining dataset, enabling direct comparison of voxel-level brain age prediction, image inpainting, and the proposed multitask framework under the same pretraining-data configuration used for ischemic stroke lesion segmentation.

A fixed validation set of 15 scans and a fixed test set of 20 scans were used throughout all experiments to ensure direct comparability between the evaluated methods. From the remaining scans, nested training subsets containing 10, 25, 45, and 65 cases were constructed such that each smaller subset was fully contained within the subsequent larger subset. Identical training subsets were used across all training strategies at each dataset size to ensure that differences in segmentation performance were attributable to the learning strategy rather than differences in the sampled training cohorts. The progressively increasing training subsets enabled a systematic evaluation of the effectiveness of pretraining under different levels of labeled data availability and allowed us to investigate whether the relative benefit of the different pretraining strategies changed as additional annotated data became available.

The fine-tuning strategies evaluated for cortical brain structure segmentation followed those identified as the most effective in the preceding downstream experiments. Specifically, the voxel-level brain age prediction and image inpainting pretrained models were fine-tuned by updating all network parameters. The proposed multitask pretrained model was evaluated using two fine-tuning strategies: fine-tuning all pretrained network parameters and randomly reinitializing the decoder while retaining the pretrained encoder weights before fine-tuning the entire network. A model trained from scratch using the same network architecture was included as the baseline to provide a direct comparison with the pretrained models.

All models were trained for 300 epochs using the Adam optimizer with an initial learning rate of $1\times10^{-3}$, a weight decay of $1\times10^{-4}$, and a StepLR learning-rate scheduler with a step size of 150 epochs. For the model pretrained using voxel-level brain age prediction, an additional learning-rate scheduling configuration with a reduced step size of 100 epochs was evaluated to investigate whether more frequent learning-rate decay could improve downstream fine-tuning performance. However, the original configuration with a step size of 150 epochs consistently provided better performance and was therefore retained for the final comparison and reported results. An effective batch size of 8 was used for all experiments. The model achieving the lowest validation loss was selected for evaluation on the test set. Each experimental configuration, defined by the combination of training-set size and training strategy, was repeated five times using different random seeds to account for stochastic variation during training. All reported quantitative results are presented as the mean $\pm$ standard deviation across the five independent runs.



\section{Results} \label{sec:results}

\subsection{Self-Supervised Pretraining Strategies}
\begin{table}[ht]
\centering
\caption{Backbone architecture comparison for the voxel-level brain age prediction pretext task. Lower MAE indicates better performance.}
\label{tab:architecture}

\begin{tabular}{lcc}
\hline
\textbf{Architecture} &
\textbf{Multi-source T1w} &
\textbf{CNS (T1w+FLAIR)} \\
&
\textbf{MAE (years)} &
\textbf{MAE (years)} \\
\hline
U-Net      & $6.4 \pm 4.3$ & $6.9 \pm 4.7$ \\
UNETR      & $7.1 \pm 4.4$ & $8.0 \pm 5.0$ \\
SwinUNETR  & \textbf{6.1 $\pm$ 4.1} & \textbf{5.3 $\pm$ 3.1} \\
\hline
\end{tabular}

\end{table}

To identify the most suitable backbone architecture for the proposed voxel-level brain age prediction framework, an ablation study was conducted using three state-of-the-art architectures: U-Net, UNETR, and SwinUNETR. The comparison was performed under two pretraining settings: (1) using T1w MR from the multi-source public dataset, and (2) using concatenated T1w and FLAIR MR from CNS dataset. As shown in Table~\ref{tab:architecture}, SwinUNETR consistently achieved the lowest MAE in both settings and was therefore selected as the backbone architecture for all subsequent pretraining and downstream segmentation experiments.
\\
Table~\ref{tab:pretext_results} summarizes the performance of the three self-supervised pretraining strategies under the two pretraining settings. Compared with the corresponding single-task frameworks, the proposed multitask model achieved improved reconstruction performance and voxel-level brain age prediction accuracy. This suggests that the two pretext tasks provide complementary supervisory signals, motivating the evaluation of whether the jointly learned representations improve performance on downstream segmentation tasks.

\begin{table*}[ht]
\centering
\caption{Performance of the three self-supervised pretraining strategies under the two pretraining settings. Lower perceptual loss and voxel MAE indicate better performance, whereas higher PSNR indicates better reconstruction quality.}
\label{tab:pretext_results}

\begin{tabular}{ll|ccc}
\hline
\textbf{Pretraining Strategy} &
\textbf{} &
\textbf{Perceptual Loss} &
\textbf{PSNR (dB)} &
\textbf{Voxel MAE (years)} \\
\hline

\multirow{2}{*}{Image Inpainting}
& Multi-source T1w
& 0.0240
& 34.90
& -- \\

& CNS (T1w+FLAIR)
& 0.0033
& 43.52
& -- \\

\hline

\multirow{2}{*}{Voxel-Level Brain Age}
& Multi-source T1w
& --
& --
& $6.1 \pm 4.1$ \\

& CNS (T1w+FLAIR)
& --
& --
& \textbf{$5.3 \pm 3.1$} \\

\hline

\multirow{2}{*}{Multitask}
& Multi-source T1w
& \textbf{0.0020}
& \textbf{46.24}
& \textbf{5.46 $\pm$ 3.7} \\

& CNS (T1w+FLAIR)
& \textbf{0.0026}
& \textbf{45.78}
& \textbf{5.25 $\pm$ 3.1} \\

\hline
\end{tabular}
\end{table*}

\subsection{Fine-tuning on Downstream Segmentation Tasks}

\subsubsection{Multiple Sclerosis Lesion Segmentation}

\begin{table*}[ht]
\centering
\caption{Comparison of different fine-tuning strategies for MS lesion segmentation. For all pretrained models, the segmentation decoder was reinitialized prior to fine-tuning. Each experiment was independently repeated five times, and the mean Dice score $\pm$ standard deviation is reported. The best result for each training set is highlighted in \textbf{\textit{bold italic}}, while \textbf{bold} indicates results within a Dice difference of 0.005 from the best-performing method.}
\label{tab:ms_finetuning}
\scriptsize
\resizebox{\textwidth}{!}{
\begin{tabular}{c|c|cc|cc|cc}
\hline
\textbf{Training} &
\textbf{Scratch} &
\multicolumn{2}{c|}{\textbf{Voxel-Level Brain Age}} &
\multicolumn{2}{c|}{\textbf{Image Inpainting}} &
\multicolumn{2}{c}{\textbf{Multitask}} \\

\textbf{Size} &
\textbf{End-to-End} &
\textbf{All Layers} &
\textbf{Decoder Only} &
\textbf{All Layers} &
\textbf{Decoder Only} &
\textbf{All Layers} &
\textbf{Decoder Only} \\
\hline

11 &
0.662 $\pm$ 0.016 &
0.631 $\pm$ 0.013 &
0.593 $\pm$ 0.006 &
0.654 $\pm$ 0.022 &
0.623 $\pm$ 0.008 &
\textbf{\textit{0.671 $\pm$ 0.016}} &
0.654 $\pm$ 0.014 \\

22 &
\textbf{\textit{0.699 $\pm$ 0.015}} &
0.692 $\pm$ 0.012 &
0.685 $\pm$ 0.017 &
\textbf{0.696 $\pm$ 0.019} &
0.661 $\pm$ 0.009 &
\textbf{0.695 $\pm$ 0.025} &
0.699 $\pm$ 0.010 \\

30 &
\textbf{\textit{0.751 $\pm$ 0.013}} &
0.715 $\pm$ 0.015 &
0.722 $\pm$ 0.010 &
0.727 $\pm$ 0.012 &
0.700 $\pm$ 0.009 &
0.737 $\pm$ 0.006 &
0.706 $\pm$ 0.002 \\

42 &
\textbf{\textit{0.757 $\pm$ 0.017}} &
0.733 $\pm$ 0.014 &
0.724 $\pm$ 0.014 &
0.743 $\pm$ 0.010 &
0.725 $\pm$ 0.013 &
\textbf{0.752 $\pm$ 0.012} &
0.727 $\pm$ 0.005 \\

50 &
0.757 $\pm$ 0.017 &
0.749 $\pm$ 0.011 &
0.736 $\pm$ 0.014 &
0.747 $\pm$ 0.012 &
0.724 $\pm$ 0.010 &
\textbf{\textit{0.764 $\pm$ 0.010}} &
0.741 $\pm$ 0.009 \\

58 &
0.764 $\pm$ 0.019 &
0.763 $\pm$ 0.008 &
0.750 $\pm$ 0.010 &
0.765 $\pm$ 0.008 &
0.732 $\pm$ 0.014 &
\textbf{\textit{0.775 $\pm$ 0.006}} &
0.751 $\pm$ 0.002 \\

\hline
\end{tabular}
}
\end{table*}

\begin{figure*}[H]
    \centering
    \includegraphics[width=1.7\columnwidth]{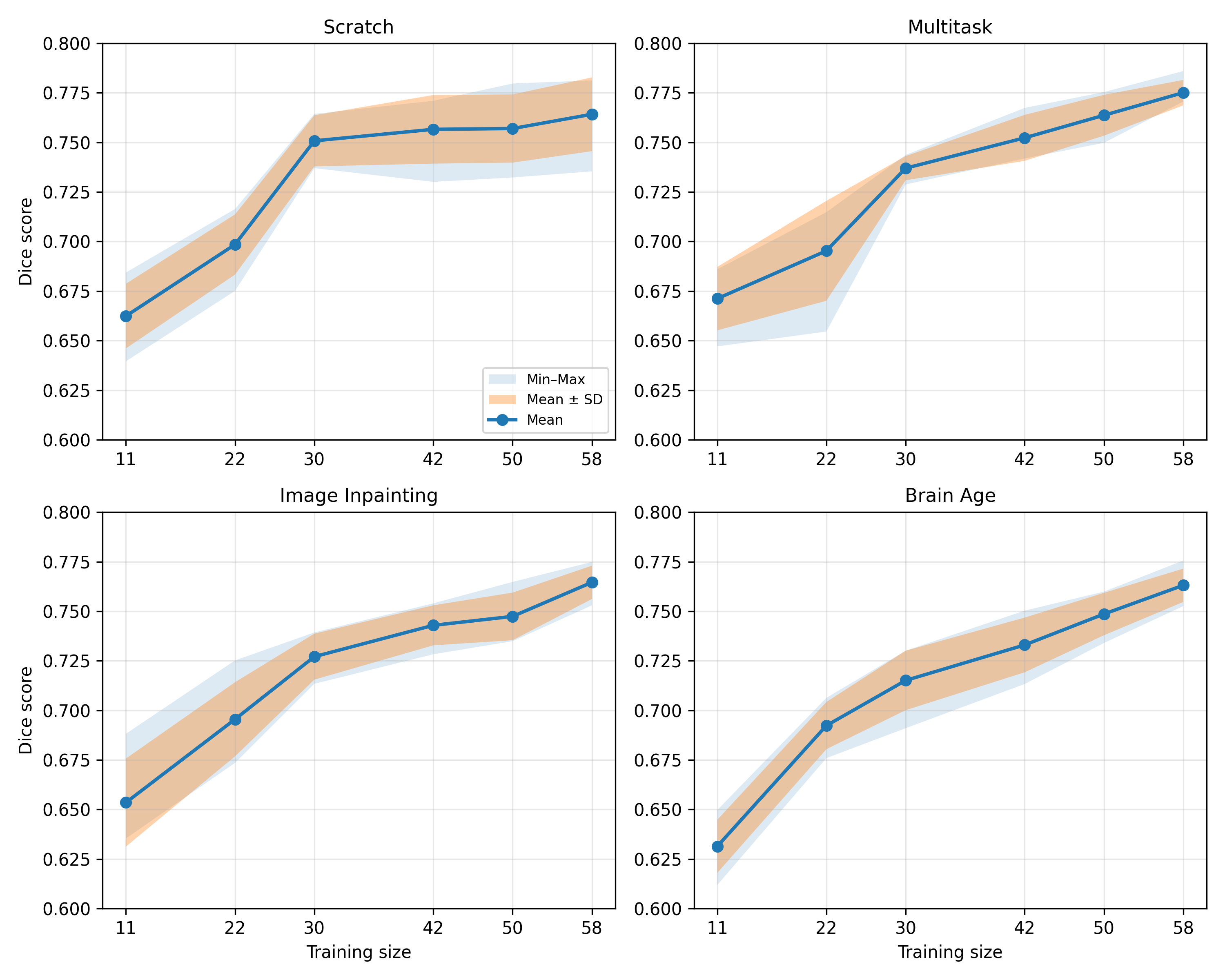}

    \caption{Comparison of the best-performing fine-tuning strategy for the MS lesion segmentation task, in which the entire network was fine-tuned after reinitializing the segmentation decoder.}
    \label{fig:performance}
\end{figure*}
\begin{figure*}[htbp]
    \centering
    \includegraphics[width=\textwidth]{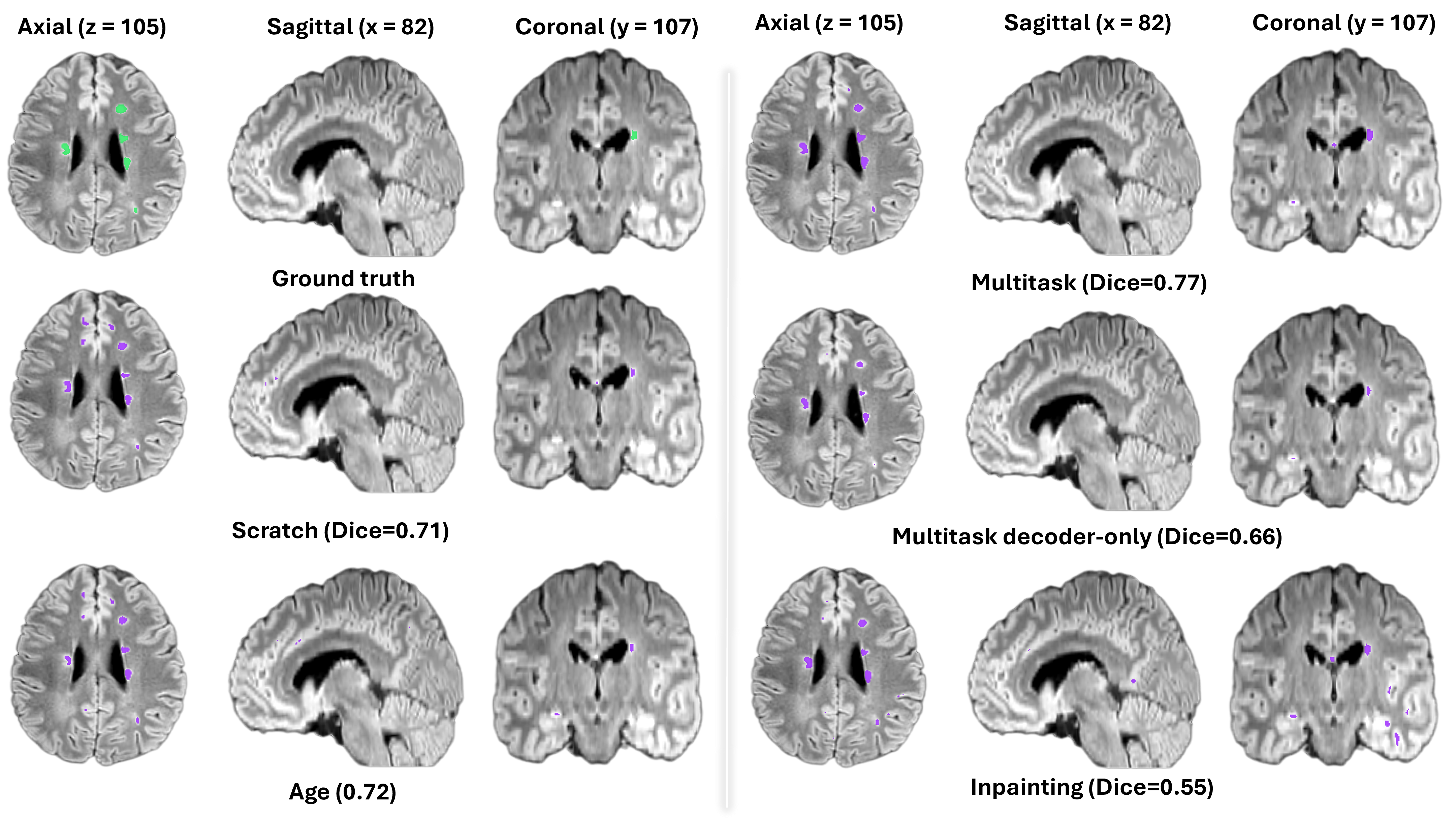}
    \caption{
    MS lesion segmentation results for the same representative test image across the evaluated pretraining strategies with 11 images used for fine-tuning. Although both T1w and FLAIR images were used as model inputs, the segmentation results are overlaid on the FLAIR image for clearer visualization of the lesions. For all pretrained models, the decoder was reinitialized before fine-tuning; all layers were subsequently fine-tuned except for the Multitask decoder-only setting, in which only the reinitialized decoder was trained.
    }
    \label{fig:ms_lesion_segmentation_11}
\end{figure*}

To determine the most effective transfer learning strategy, multiple fine-tuning configurations were initially evaluated for each pretrained model, together with a model trained from scratch for comparison. Based on these experiments, fine-tuning the entire network while reinitializing the segmentation decoder consistently achieved the best performance for the pretrained models and was therefore selected for the final comparison. Table~\ref{tab:ms_finetuning} summarizes the quantitative Dice scores obtained using the best-performing fine-tuning strategy, while Figure~\ref{fig:performance} illustrates the corresponding performance variability across the different training set sizes.

Among the evaluated pretraining strategies, the proposed multitask framework generally achieved the highest Dice scores, outperforming both image inpainting and voxel-level brain age prediction across most training set sizes. Compared with training from scratch, multitask pretraining provided the largest improvements when only a very limited number of annotated training samples were available (11 subjects) and continued to outperform training from scratch at the larger training set sizes (50 and 58 subjects). For the 22- and 42-subject training sets, the two approaches achieved comparable performance, whereas training from scratch produced the highest Dice score for the 30-subject training set. Although the available dataset does not allow us to determine whether this trend would persist with substantially larger training sets, these findings suggest that the proposed multitask pretraining is particularly beneficial in low-data settings while remaining competitive as more annotated training data become available. Qualitative results for a representative test case in the 11-subject fine-tuning setting are shown in Fig.~\ref{fig:ms_lesion_segmentation_11}.
This motivated us to further evaluate the proposed framework on ischemic stroke lesion segmentation, which requires only T1w MR.

\subsubsection{Ischemic Stroke Lesion Segmentation}






\begin{figure*}[htbp]
    \centering
    \includegraphics[height=0.57\textheight, keepaspectratio]{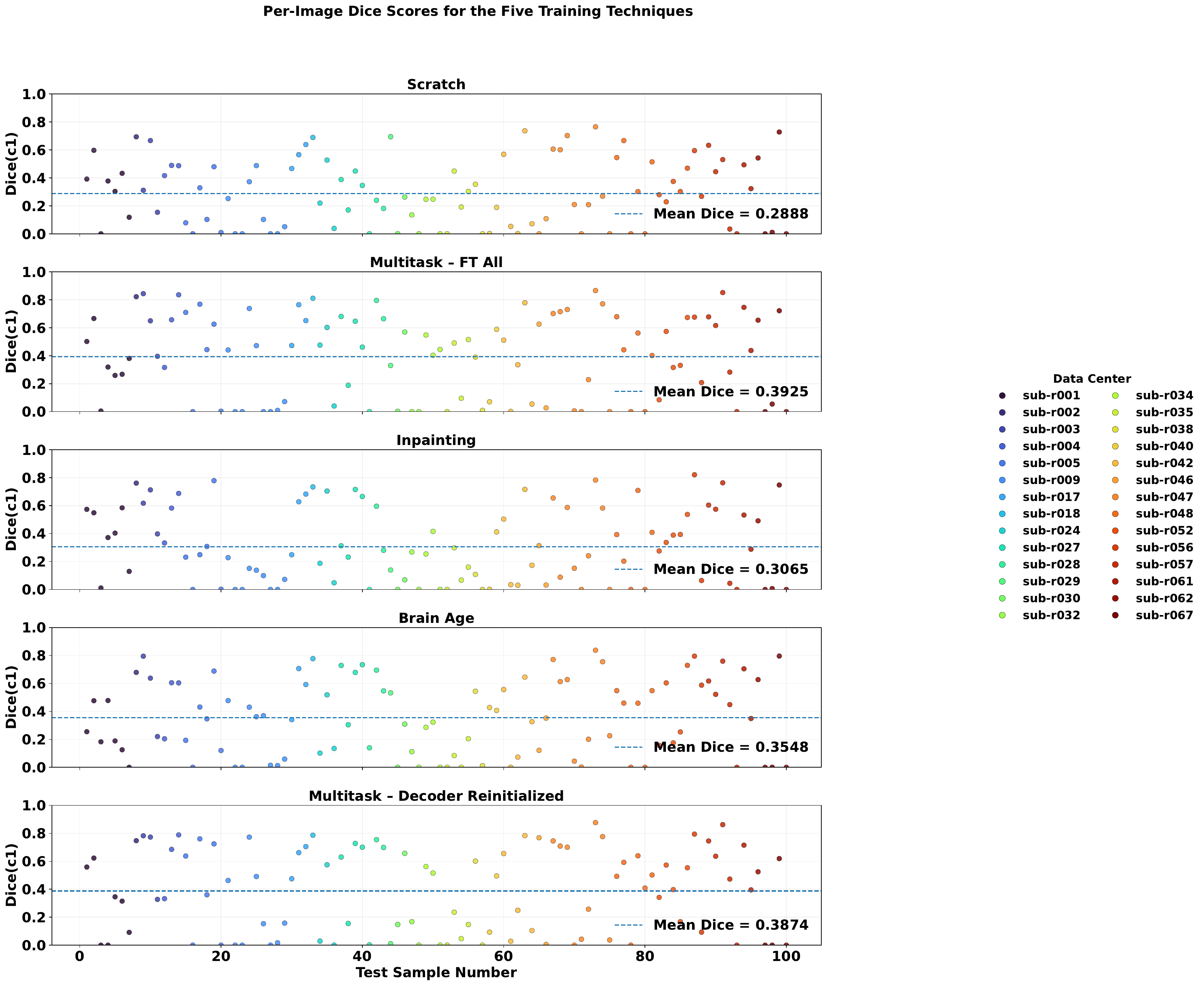}  
    \caption{Per-image Dice scores for the five training strategies on the ischemic stroke lesion segmentation task. Results are shown for the fourth independent run using 50 labeled training cases. Each point represents one of the 100 test images, displayed in the same order across all subplots to enable direct comparison between methods. Colors indicate the acquisition center of each test image, and the dashed horizontal line denotes the mean Dice score. The evaluated strategies are training from scratch, image inpainting pretraining, voxel-level brain age prediction pretraining, multitask pretraining with fine-tuning of all pretrained layers (Multitask–FT All), and multitask pretraining with decoder reinitialization followed by end-to-end fine-tuning (Multitask–Decoder Reinitialized).}
    \label{fig:strokeanalysis}
\end{figure*}

\begin{figure*}[htbp]
    \centering
    \includegraphics[width=\textwidth]{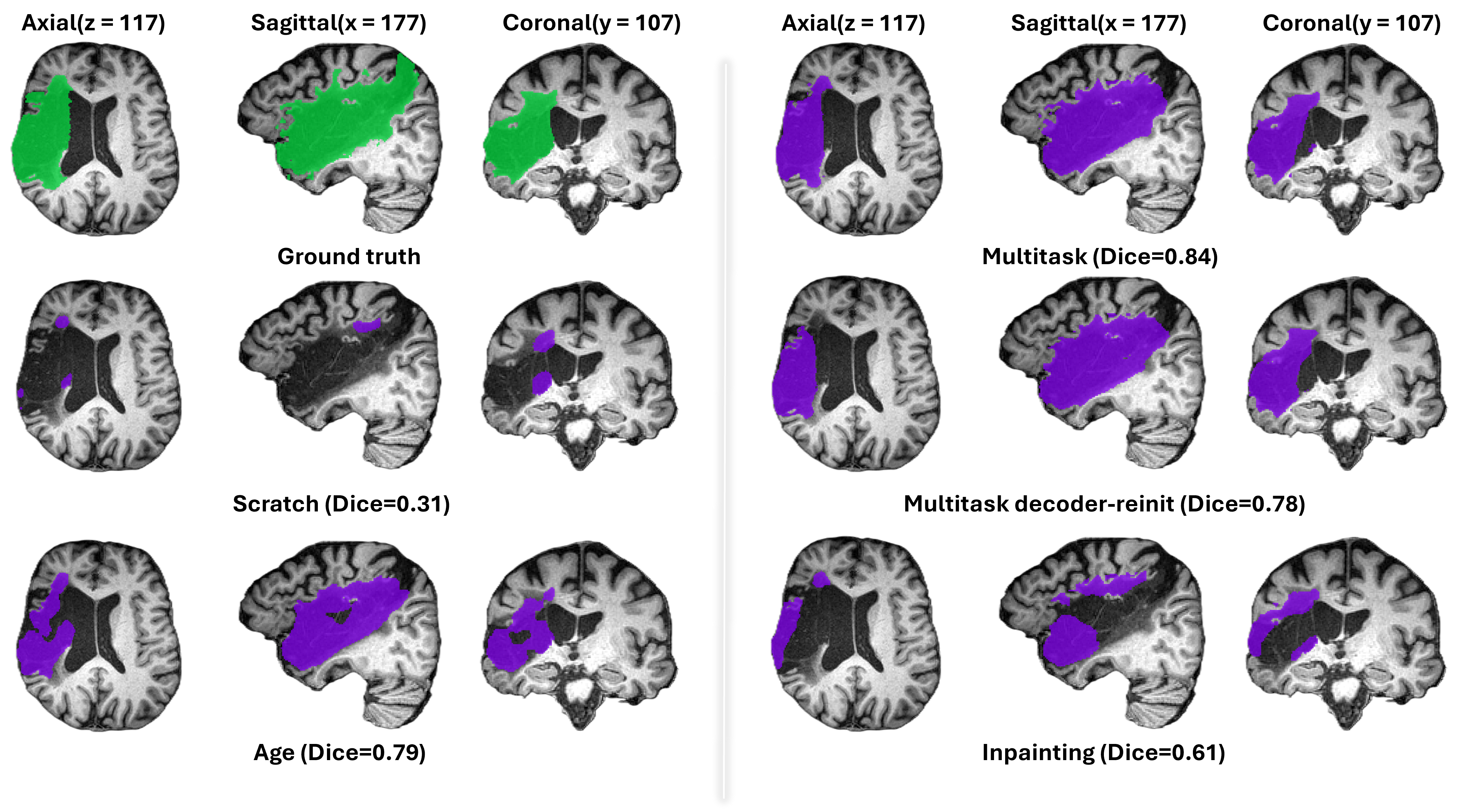}
    \caption{
    Stroke lesion segmentation results for the same representative test image across the evaluated pretraining strategies with 50 images used for fine-tuning in a single run. The segmentation results are overlaid on the T1w image, with the ground-truth lesion shown in green and the predicted lesions shown in violet.
    }
    \label{fig:stroke_lesion_segmentation_50}
\end{figure*}
\begin{table*}[htbp]
\centering
\caption{Mean Dice score $\pm$ standard deviation over five independent runs for ischemic stroke lesion segmentation across different training-set sizes and training strategies.}
\label{tab:stroke_dice_results}

\begin{tabular}{c c c c c c}
\toprule
\textbf{Training Size}
& \textbf{Scratch}
& \textbf{Brain Age}
& \textbf{Inpainting}
& \textbf{Multitask--Decoder Reinitialized}
& \textbf{Multitask--FT All} \\
\midrule

10
& 0.216 $\pm$ 0.033
& 0.203 $\pm$ 0.010
& 0.206 $\pm$ 0.028
& 0.276 $\pm$ 0.027
& \textbf{0.286 $\pm$ 0.011} \\
20
& 0.229 $\pm$ 0.010
& 0.249 $\pm$ 0.010
& 0.231 $\pm$ 0.024
& 0.283 $\pm$ 0.030
& \textbf{0.300 $\pm$ 0.030} \\
50
& 0.332 $\pm$ 0.025
& 0.351 $\pm$ 0.016
& 0.329 $\pm$ 0.023
& \textbf{0.399 $\pm$ 0.016}
& 0.391 $\pm$ 0.009 \\
100
& 0.459 $\pm$ 0.017
& 0.459 $\pm$ 0.009
& 0.464 $\pm$ 0.016
& \textbf{0.480 $\pm$ 0.014}
& 0.470 $\pm$ 0.012 \\
150
& 0.488 $\pm$ 0.009
& 0.485 $\pm$ 0.009
& 0.488 $\pm$ 0.015
& 0.481 $\pm$ 0.024
& \textbf{0.492 $\pm$ 0.011} \\
200
& 0.506 $\pm$ 0.010
& 0.506 $\pm$ 0.008
& 0.507 $\pm$ 0.022
& \textbf{0.525 $\pm$ 0.008}
& 0.502 $\pm$ 0.031 \\
250
& 0.526 $\pm$ 0.004
& 0.517 $\pm$ 0.019
& 0.502 $\pm$ 0.017
& 0.522 $\pm$ 0.008
& \textbf{0.536 $\pm$ 0.001} \\
300
& 0.530 $\pm$ 0.013
& 0.533 $\pm$ 0.013
& 0.525 $\pm$ 0.013
& 0.520 $\pm$ 0.009
& \textbf{0.534 $\pm$ 0.006} \\

350
& 0.539 $\pm$ 0.010
& \textbf{0.546 $\pm$ 0.005}
& 0.524 $\pm$ 0.017
& 0.536 $\pm$ 0.011
& 0.533 $\pm$ 0.005 \\
400
& 0.537 $\pm$ 0.011
& 0.539 $\pm$ 0.008
& 0.533 $\pm$ 0.014
& \textbf{0.542 $\pm$ 0.013}
& 0.540 $\pm$ 0.009 \\

\bottomrule

\end{tabular}

\end{table*}

Table~\ref{tab:stroke_dice_results} summarizes the performance of the evaluated training strategies for ischemic stroke lesion segmentation across different training-set sizes, reporting the mean Dice score ($\pm$ standard deviation) over five independent runs for each method.

As shown in Table~\ref{tab:stroke_dice_results}, the pretrained models generally outperform the model trained from scratch in data-scarce scenarios. The proposed multitask pretraining framework consistently achieves the highest segmentation performance for training-set sizes up to 150 labeled cases. Although the relative ranking of the two multitask fine-tuning strategies varies slightly across individual training-set sizes, both consistently outperform the individual pretraining strategies and the randomly initialized baseline. Specifically, fine-tuning all pretrained layers provides the highest performance for the 10- and 20-case settings, whereas decoder reinitialization followed by end-to-end fine-tuning yields the best performance for the 50- and 100-case settings. At 150 labeled cases, the two multitask strategies achieve comparable Dice scores, indicating that no single fine-tuning strategy is uniformly optimal across the low-data regime.

The results in Table~\ref{tab:stroke_dice_results} also demonstrate that the benefit of pretraining is most pronounced when labeled data are scarce. The proposed multitask framework provides the largest performance gains over training from scratch, reaching improvements of approximately 6--7 Dice percentage points for the smallest training subsets. As the number of labeled training cases increases, the performance gap between the pretrained and randomly initialized models progressively decreases, becoming minimal beyond approximately 200 training cases. This trend indicates that pretraining is particularly beneficial in the low-data regime, whereas its relative advantage diminishes as more labeled data become available for supervised training.

Another notable observation is the difference in performance stability among the evaluated pretraining strategies. Beyond approximately 200 labeled training cases, the image-inpainting approach exhibits larger fluctuations in mean Dice across consecutive training-set sizes, suggesting greater sensitivity to the amount or composition of the labeled training data. In contrast, both multitask variants show more stable performance as the training-set size increases. Among them, fine-tuning all pretrained layers exhibits the most consistent behavior across the evaluated training-set sizes, whereas decoder reinitialization shows somewhat greater variability while remaining more stable than the individual pretraining approaches.
To further examine performance at the individual-case level, Fig.~\ref{fig:strokeanalysis} shows the per-image Dice scores for all 100 test cases from the fourth independent run using 50 labeled training cases. The figure highlights the substantial inter-case variability in segmentation performance while enabling direct comparison of the evaluated strategies on the same test images. A qualitative comparison for a representative test case from the same 50-case fine-tuning setting is provided in Fig.~\ref{fig:stroke_lesion_segmentation_50}. The same test image and anatomical slices are shown across all evaluated strategies, with the ground-truth lesion in green and the corresponding predictions in violet, enabling direct visual comparison of lesion localization and extent.

Overall, these findings indicate that combining voxel-level brain age prediction and image inpainting within a unified multitask pretraining framework yields representations that transfer more effectively to downstream ischemic stroke lesion segmentation than either pretext task individually. The advantage of multitask pretraining is particularly evident in the low-data regime, while the differences among training strategies become progressively smaller as the amount of labeled training data increases.

\subsubsection{Cortical Brain Structure segmentation}
\begin{table*}[htbp]
\centering
\caption{Mean Dice score $\pm$ standard deviation over five independent runs for cortical brain structure segmentation across different training-set sizes and training strategies.}
\label{tab:cortical_dice_results}

\begin{tabular}{c c c c c c}
\toprule
\textbf{Training Size}
& \textbf{Scratch}
& \textbf{Brain Age}
& \textbf{Inpainting}
& \textbf{Multitask--FT All}
& \textbf{Multitask--Decoder Reinitialized} \\
\midrule

10
& 0.186 $\pm$ 0.023
& 0.063 $\pm$ 0.010
& 0.221 $\pm$ 0.024
& \textbf{0.248 $\pm$ 0.033}
& \textbf{0.251 $\pm$ 0.049} \\

25
& 0.493 $\pm$ 0.063
& 0.313 $\pm$ 0.047
& 0.545 $\pm$ 0.080
& \textbf{0.633 $\pm$ 0.044}
& 0.611 $\pm$ 0.033 \\

45
& 0.726 $\pm$ 0.035
& 0.723 $\pm$ 0.019
& \textbf{0.762 $\pm$ 0.005}
& 0.732 $\pm$ 0.041
& 0.674 $\pm$ 0.060 \\

65
& 0.741 $\pm$ 0.016
& 0.737 $\pm$ 0.036
& \textbf{0.772 $\pm$ 0.015}
& 0.762 $\pm$ 0.016
& 0.742 $\pm$ 0.033 \\

\bottomrule
\end{tabular}

\end{table*}
\
\begin{figure*}[htbp]
    \centering
    \includegraphics[height=0.4\textheight, keepaspectratio]{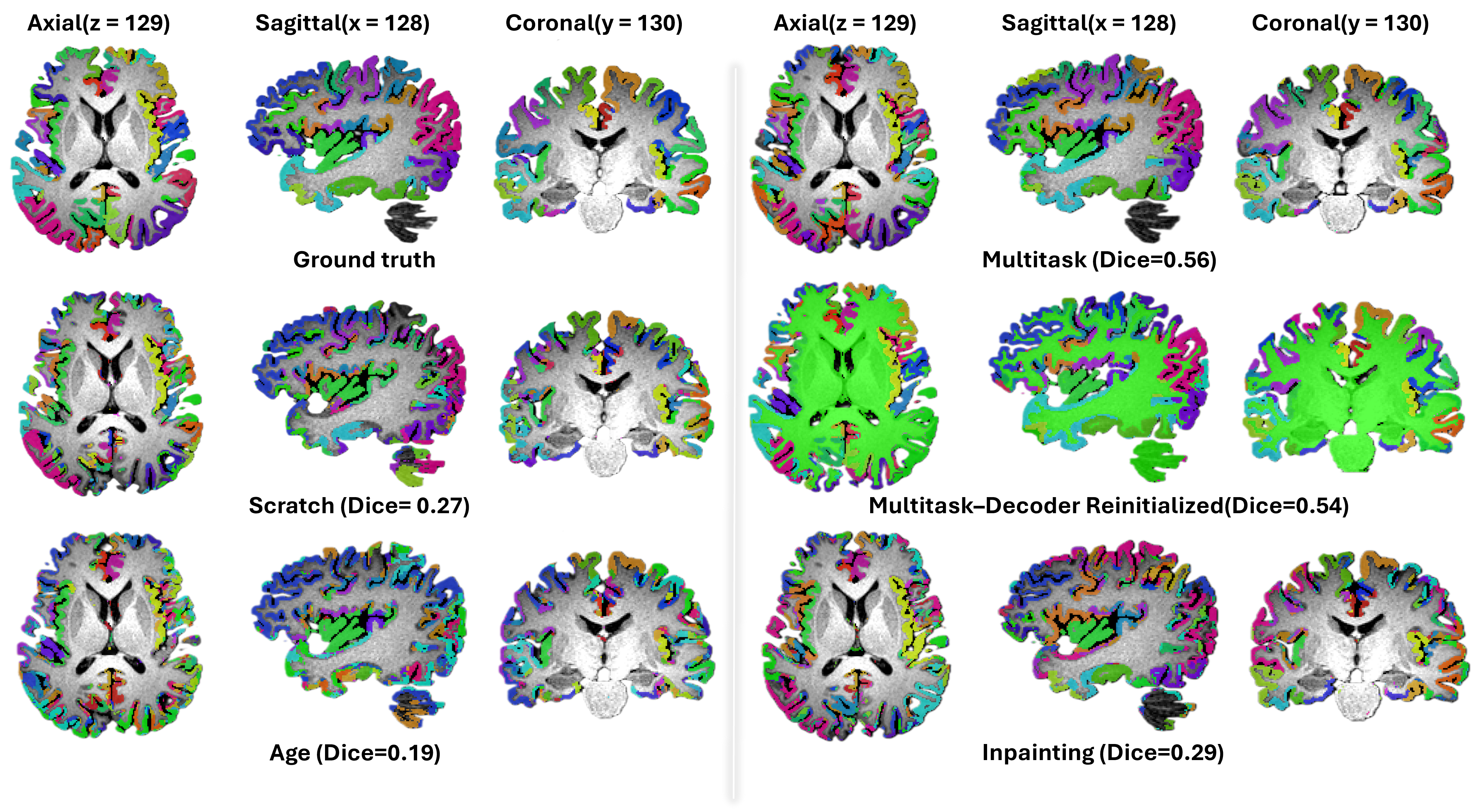}
    \caption{
    Cortical parcellation results for the same representative test image across the evaluated pretraining strategies with 25 images used for fine-tuning.
    }
    \label{fig:cortical_segmentation_25_run4}
\end{figure*}

Table \ref{tab:cortical_dice_results} summarizes the cortical brain structure segmentation performance across the four training-set sizes. The proposed multitask pretraining framework outperformed training from scratch across all evaluated training-set sizes, with the largest improvements observed in the low-data regime. At 10 training cases, Multitask--Decoder Reinitialized achieved the highest mean Dice score of $0.251 \pm 0.049$, compared with $0.186 \pm 0.023$ for training from scratch. Multitask--FT All achieved a similar Dice score of $0.248 \pm 0.033$. At 25 training cases, Multitask--FT All achieved the highest performance ($0.633 \pm 0.044$), followed by Multitask--Decoder Reinitialized ($0.611 \pm 0.033$), compared with $0.493 \pm 0.063$ for training from scratch. A representative qualitative example from the same 25-case fine-tuning setting is shown in Fig.~\ref{fig:cortical_segmentation_25_run4}. The same test image slices are displayed across the evaluated strategies, enabling direct visual comparison of the predicted cortical regions.

As the number of labeled training cases increased, the relative performance of the evaluated pretraining strategies changed. At 45 and 65 training cases, image inpainting achieved the highest mean Dice scores, reaching $0.762 \pm 0.005$ and $0.772 \pm 0.015$, respectively. At these training-set sizes, Multitask--FT All remained competitive, achieving Dice scores of $0.732 \pm 0.041$ and $0.762 \pm 0.016$, respectively, and continued to outperform the corresponding models trained from scratch.

In contrast, voxel-level brain age prediction alone did not provide a consistent advantage over random initialization for cortical brain structure segmentation. The brain age pretrained model achieved lower mean Dice scores than training from scratch at all four training-set sizes, with the difference being particularly pronounced in the low-data settings. At 10 training cases, brain age pretraining achieved a mean Dice score of $0.063 \pm 0.010$, compared with $0.186 \pm 0.023$ for training from scratch. Similarly, at 25 cases, the corresponding Dice scores were $0.313 \pm 0.047$ and $0.493 \pm 0.063$, respectively. The difference became substantially smaller as the training-set size increased, with brain age pretraining and training from scratch achieving comparable performance at 45 and 65 cases.
Overall, multitask pretraining achieved the highest performance in the low-data settings, whereas image inpainting performed best with the larger training subsets, while voxel-level brain age pretraining alone did not outperform training from scratch.

\section{Discussion and Future Perspectives}

In this study, we investigated voxel-level brain age prediction as a domain-specific self-supervised pretext task and compared it with image inpainting, a widely used non-domain-specific self-supervised objective. Building upon these complementary pretext tasks, we proposed a multitask self-supervised learning framework that jointly optimizes both objectives to learn richer and more transferable neuroimaging representations. Rather than optimizing the learned representations for a single downstream application, our objective was to develop a pretraining strategy capable of generalizing across diverse neurological tasks, regardless of their direct relationship to brain aging.

To evaluate this hypothesis, we considered three downstream segmentation tasks with distinct characteristics: cortical brain structure segmentation, representing an anatomical task involving structures known to be affected by aging, and MS and ischemic stroke lesion segmentation, representing two clinically relevant pathological tasks not directly related to the normal brain aging process. This experimental design enabled us to investigate whether representations learned through voxel-level brain age prediction preferentially benefit anatomically age-sensitive applications, whether the more generic image inpainting objective provides broader transferability, or whether jointly optimizing both objectives produces complementary representations that are particularly advantageous when labeled downstream data are scarce.

Overall, the proposed multitask framework achieved the strongest or among the strongest performance across the evaluated downstream tasks, with its advantage being particularly evident in several low-data settings. These findings suggest that combining domain-specific and non-domain-specific self-supervision can provide more transferable representations than relying exclusively on either objective.

Unlike the other downstream tasks, MS lesion segmentation requires both T1w and FLAIR MR images because MS lesions are often difficult to identify on T1w images alone and are more conspicuous on FLAIR. Consequently, pretraining for this task was restricted to the CNS dataset, which provides paired T1w and FLAIR images together with chronological age information. Neither image inpainting nor voxel-level brain age prediction alone consistently outperformed training from scratch, whereas their combination within the proposed multitask framework generally yielded the strongest performance.

One factor specific to the MS experiments is the limited diversity of the available pretraining data. Because paired T1w and FLAIR images were required, pretraining was restricted to the single-source CNS cohort. Although this ensured consistency between the pretraining and downstream input modalities, the limited variation in imaging protocols, scanners, and subject populations may have constrained the transferability of the learned representations. A larger multi-source normative dataset containing paired T1w and FLAIR images together with chronological age information would enable a more comprehensive evaluation of this hypothesis and help distinguish limitations associated with the pretext tasks themselves from those arising from the scale and diversity of the available pretraining data. 

Although multitask pretraining generally outperformed training from scratch in the MS experiments, this trend was not observed for the training subset containing 30 subjects, where the randomly initialized model achieved the highest Dice score. Given the relatively small size of the MS lesion segmentation dataset, it remains unclear whether this result reflects statistical variability associated with the sampled training data or a genuine characteristic of the learning process. This observation motivated the evaluation of the proposed framework on an additional non-age-related pathological task, ischemic stroke lesion segmentation. 

In contrast to the MS experiments, ischemic stroke lesion segmentation enabled all pretraining strategies to use the larger multi-source T1w normative dataset. This substantially increased the diversity of the pretraining data by incorporating scans from multiple publicly available cohorts with differences in imaging protocols, scanners, and subject populations. The stroke experiments therefore provided an opportunity to evaluate the proposed framework under a substantially broader pretraining setting.

The stroke lesion segmentation results demonstrated a clearer advantage of self-supervised pretraining than that observed for MS lesion segmentation. The benefit was most pronounced when labeled downstream data were limited, although its magnitude varied among the evaluated strategies. The proposed multitask framework achieved the highest segmentation performance for training-set sizes up to 150 labeled cases. Fine-tuning all pretrained layers yielded the highest Dice scores with 10 and 20 training cases, whereas decoder reinitialization followed by end-to-end fine-tuning performed best with 50 and 100 cases. At 150 cases, the two multitask fine-tuning strategies achieved comparable performance. The largest gains over training from scratch reached approximately 6--7 absolute Dice percentage points in the smallest training subsets. As the number of labeled cases increased, however, the performance difference between pretrained and randomly initialized models progressively diminished. This pattern indicates that the value of the learned initialization is greatest when downstream supervision is limited and becomes less pronounced as sufficient task-specific annotations become available. 

Differences in performance stability were also observed among the individual pretraining strategies. Image inpainting exhibited greater variation as the number of labeled training cases increased, suggesting greater sensitivity to the composition of the downstream training set. Voxel-level brain age prediction generally showed more stable behavior, although it did not consistently outperform training from scratch. The multitask framework, in comparison, combined strong performance with relatively stable behavior across the evaluated training-set sizes. 

The per-image analysis further demonstrated that the improvements obtained with multitask pretraining were observed across images acquired from multiple centers rather than being driven by a small subset of test cases. Higher Dice scores were observed across a broad range of subjects and acquisition centers, providing additional evidence that the improvements were not restricted to a particular center or imaging condition.

The cortical brain structure segmentation experiments revealed a different pattern from the lesion segmentation tasks. Multitask pretraining achieved the strongest performance in the low-data settings of 10 and 25 training cases, whereas image inpainting achieved the highest performance with 45 and 65 labeled cases. Unexpectedly, voxel-level brain age prediction alone did not outperform training from scratch at any of the evaluated training-set sizes, despite cortical anatomy being affected by the aging process. 

One possible explanation is that, although cortical morphology undergoes age-related changes, cortical parcellation is not itself a directly age-prediction-related task. Cortical parcellation primarily requires the identification of anatomical boundaries and spatial relationships among neighboring cortical regions, whereas voxel-level brain age prediction encourages the network to identify features associated with normative aging trajectories. Features that are informative for estimating chronological age may therefore not necessarily be those that are most discriminative for delineating cortical regions. 

The stronger performance of image inpainting with the larger cortical training subsets may also reflect the nature of the downstream task. Image inpainting requires reconstruction of missing image regions from their surrounding anatomical context, encouraging the model to learn local structural information, spatial relationships, and anatomical organization. Such representations may be particularly relevant to cortical parcellation, where accurate segmentation depends heavily on distinguishing neighboring anatomical structures. As additional labeled downstream data became available, these generic structural representations may have provided a more suitable initialization for supervised cortical parcellation than representations optimized specifically for brain age prediction. 

Nevertheless, the cortical experiments also provide an important observation regarding the proposed multitask formulation. Despite the limited performance of voxel-level brain age prediction when used independently, combining it with image inpainting substantially improved performance in the smallest training subsets. At 10 and 25 training cases, both multitask variants outperformed training from scratch and the individual pretraining strategies. The utility of the brain age objective should therefore not be assessed exclusively from its performance as an isolated pretext task. Within the joint optimization framework, age-related supervision may complement the structural and contextual information learned through image reconstruction, resulting in a more informative representation when downstream supervision is particularly limited. 

Taken together, the three downstream experiments indicate that the effectiveness of self-supervised pretraining depends on the interaction among the pretext objective, downstream task, amount of labeled downstream data, and diversity of the pretraining cohort. No individual pretext task was uniformly optimal across all experimental conditions. In contrast, the proposed multitask framework demonstrated a particular advantage in several low-data settings, supporting the central motivation of combining domain-specific age-related supervision with a generic reconstruction objective to capture complementary aspects of brain MR images.

Several limitations should be considered when interpreting these findings. First, identifying an appropriate publicly available dataset with high-quality manual annotations for a downstream task directly related to brain aging was challenging. Although cortical anatomy is affected by aging, cortical parcellation primarily evaluates anatomical boundary delineation and therefore represents only an indirect assessment of the transferability of age-related representations. Future work should evaluate the proposed framework on downstream tasks more directly associated with aging and age-related structural changes.

Second, the present evaluation was restricted to segmentation tasks. Although the three selected applications provide substantially different segmentation settings, including healthy anatomy and two pathological lesion segmentation problems, they do not fully establish the generalizability of the learned representations across other types of neuroimaging applications. Future work should therefore extend the evaluation to classification and prediction tasks, including neurological disease classification, cognitive or clinical outcome prediction, and other applications in which age-related representations may provide complementary information.

Finally, the MS experiments were constrained by the availability of multimodal normative pretraining data. Because MS lesion segmentation required paired T1w and FLAIR images, pretraining was restricted to a single-source cohort, whereas the T1w-only stroke and cortical experiments benefited from a larger multi-source pretraining dataset. Future work should investigate large-scale, multi-center normative datasets containing paired T1w and FLAIR images and chronological age information to determine whether increasing pretraining diversity yields similar improvements for multimodal downstream tasks.

\section{Conclusion}

This study investigated voxel-level brain age prediction as a domain-specific self-supervised pretext task and introduced a multitask framework that combines it with image inpainting to learn more transferable brain MR imaging representations. Across three downstream segmentation tasks, the proposed multitask approach demonstrated its greatest benefit in low-data settings, highlighting the value of combining complementary domain-specific and generic supervisory signals when annotated data are limited. The findings also indicate that the effectiveness of self-supervised pretraining is influenced by the scale and diversity of the pretraining data, with stronger transfer observed when a larger multi-source pretraining cohort was available. Overall, these results support multitask self-supervised learning as a promising approach for improving representation transfer in brain MR imaging, particularly in data scarce scenarios.

\section{Acknowledgments}
This work was supported by research funding from the Natural Sciences and Engineering Research Council of Canada (NSERC) and the Canada
Research Chairs Program. We also acknowledge the Digital Research Alliance of
Canada for providing computational infrastructure support.

\bibliographystyle{plainnat}

\bibliography{cas-refs}

@article{klein_101_2012,
	title = {101 Labeled Brain Images and a Consistent Human Cortical Labeling Protocol},
	volume = {6},
	issn = {1662-453X},
	url = {https://www.frontiersin.org/journals/neuroscience/articles/10.3389/fnins.2012.00171/full},
	doi = {10.3389/fnins.2012.00171},
	journaltitle = {Frontiers in Neuroscience},
	shortjournal = {Front. Neurosci.},
	publisher = {Frontiers},
	author = {Klein, Arno and Tourville, Jason},
	urldate = {2026-08-25},
	year = {2012},
}

@article{hernandez_petzsche_isles_2022,
	title = {{ISLES} 2022: A multi-center magnetic resonance imaging stroke lesion segmentation dataset},
	volume = {9},
	rights = {2022 The Author(s)},
	issn = {2052-4463},
	url = {https://www.nature.com/articles/s41597-022-01875-5},
	doi = {10.1038/s41597-022-01875-5},
	shorttitle = {{ISLES} 2022},
	pages = {762},
	number = {1},
	journaltitle = {Scientific Data},
	shortjournal = {Sci Data},
	publisher = {Nature Publishing Group},
	author = {Hernandez Petzsche, Moritz R. and de la Rosa, Ezequiel and Hanning, Uta and Wiest, Roland and Valenzuela, Waldo and Reyes, Mauricio and Meyer, Maria and Liew, Sook-Lei and Kofler, Florian and Ezhov, Ivan and Robben, David and Hutton, Alexandre and Friedrich, Tassilo and Zarth, Teresa and Bürkle, Johannes and Baran, The Anh and Menze, Björn and Broocks, Gabriel and Meyer, Lukas and Zimmer, Claus and Boeckh-Behrens, Tobias and Berndt, Maria and Ikenberg, Benno and Wiestler, Benedikt and Kirschke, Jan S.},
	urldate = {2026-08-28},
	year = {2022},
	langid = {english},
}

@article{absher_stroke_2024,
	title = {The stroke outcome optimization project: Acute ischemic strokes from a comprehensive stroke center},
	volume = {11},
	issn = {2052-4463},
	url = {https://pmc.ncbi.nlm.nih.gov/articles/PMC11297183/},
	doi = {10.1038/s41597-024-03667-5},
	shorttitle = {The stroke outcome optimization project},
	pages = {839},
	journaltitle = {Scientific Data},
	shortjournal = {Sci Data},
	author = {Absher, John and Goncher, Sarah and Newman-Norlund, Roger and Perkins, Nicholas and Yourganov, Grigori and Vargas, Jan and Sivakumar, Sanjeev and Parti, Naveen and Sternberg, Shannon and Teghipco, Alex and Gibson, Makayla and Wilson, Sarah and Bonilha, Leonardo and Rorden, Chris},
	urldate = {2026-08-28},
	year = {2024},
}

@article{cotinat_dynamics_2025,
	title = {Dynamics of Ionic and Cytotoxic Edema During Acute and Subacute Stages of Patients With Ischemic Stroke: Complementarity of 23Na {MRI} and Diffusion {MRI}},
	volume = {38},
	issn = {1099-1492},
	url = {https://onlinelibrary.wiley.com/doi/abs/10.1002/nbm.70028},
	doi = {10.1002/nbm.70028},
	shorttitle = {Dynamics of Ionic and Cytotoxic Edema During Acute and Subacute Stages of Patients With Ischemic Stroke},
	pages = {e70028},
	number = {5},
	journaltitle = {{NMR} in Biomedicine},
	author = {Cotinat, Maëva and Messaoudi, Noëlle and Robinet, Emmanuelle and Suissa, Laurent and Doche, Emilie and Guye, Maxime and Audoin, Bertrand and Bensoussan, Laurent and Ranjeva, Jean-Philippe and Zaaraoui, Wafaa},
	urldate = {2026-08-28},
	year = {2025},
	langid = {english},
}

@article{yuh_mr_1991,
	title = {{MR} imaging of cerebral ischemia: findings in the first 24 hours.},
	volume = {12},
	issn = {0195-6108},
	url = {https://pmc.ncbi.nlm.nih.gov/articles/PMC8331595/},
	shorttitle = {{MR} imaging of cerebral ischemia},
	pages = {621--629},
	number = {4},
	journaltitle = {{AJNR}: American Journal of Neuroradiology},
	shortjournal = {{AJNR} Am J Neuroradiol},
	author = {Yuh, W T and Crain, M R and Loes, D J and Greene, G M and Ryals, T J and Sato, Y},
	urldate = {2026-08-28},
	year = {1991},
}

@online{noauthor_ischemic_nodate,
	title = {Ischemic Stroke Lesion Segmentation Challenge 2026 - Grand Challenge},
	url = {https://isles-26.grand-challenge.org/dataset/},
	titleaddon = {grand-challenge.org},
	urldate = {2026-08-28},
	langid = {english},
}

@article{he_global-local_2021,
	title = {Global-local transformer for brain age estimation},
	volume = {41},
	number = {1},
	journal = {IEEE transactions on medical imaging},
	publisher = {IEEE},
	author = {He, Sheng and Grant, P Ellen and Ou, Yangming},
	year = {2021},
	pages = {213--224},
}

@inproceedings{gianchandani_multitask_2024,
	address = {Cham},
	title = {A {Multitask} {Deep} {Learning} {Model} for {Voxel}-{Level} {Brain} {Age} {Estimation}},
	isbn = {978-3-031-45676-3},
	doi = {10.1007/978-3-031-45676-3_29},
	language = {en},
	booktitle = {Machine {Learning} in {Medical} {Imaging}},
	publisher = {Springer Nature Switzerland},
	author = {Gianchandani, Neha and Ospel, Johanna and MacDonald, Ethan and Souza, Roberto},
	editor = {Cao, Xiaohuan and Xu, Xuanang and Rekik, Islem and Cui, Zhiming and Ouyang, Xi},
	year = {2024},
	pages = {283--292},
}

@article{reuter_highly_2010,
	title = {Highly accurate inverse consistent registration: A robust approach},
	volume = {53},
	issn = {1053-8119},
	url = {https://www.sciencedirect.com/science/article/pii/S1053811910009717},
	doi = {10.1016/j.neuroimage.2010.07.020},
	shorttitle = {Highly accurate inverse consistent registration},
	pages = {1181--1196},
	number = {4},
	journaltitle = {{NeuroImage}},
	shortjournal = {{NeuroImage}},
	author = {Reuter, Martin and Rosas, H. Diana and Fischl, Bruce},
	urldate = {2026-09-05},
	date = {2010-12-01},
}

@article{souza_open_2018,
	series = {Segmenting the {Brain}},
	title = {An open, multi-vendor, multi-field-strength brain {MR} dataset and analysis of publicly available skull stripping methods agreement},
	volume = {170},
	issn = {1053-8119},
	url = {https://www.sciencedirect.com/science/article/pii/S1053811917306687},
	doi = {10.1016/j.neuroimage.2017.08.021},
	urldate = {2024-12-12},
	journal = {NeuroImage},
	author = {Souza, Roberto and Lucena, Oeslle and Garrafa, Julia and Gobbi, David and Saluzzi, Marina and Appenzeller, Simone and Rittner, Letícia and Frayne, Richard and Lotufo, Roberto},
	month = apr,
	year = {2018},
	pages = {482--494},
}

@article{marcus_open_2007,
	title = {Open {Access} {Series} of {Imaging} {Studies} ({OASIS}): {Cross}-sectional {MRI} {Data} in {Young}, {Middle} {Aged}, {Nondemented}, and {Demented} {Older} {Adults}},
	volume = {19},
	issn = {0898-929X},
	shorttitle = {Open {Access} {Series} of {Imaging} {Studies} ({OASIS})},
	url = {https://doi.org/10.1162/jocn.2007.19.9.1498},
	doi = {10.1162/jocn.2007.19.9.1498},
	number = {9},
	urldate = {2024-12-12},
	journal = {Journal of Cognitive Neuroscience},
	author = {Marcus, Daniel S. and Wang, Tracy H. and Parker, Jamie and Csernansky, John G. and Morris, John C. and Buckner, Randy L.},
	month = sep,
	year = {2007},
	pages = {1498--1507},
}

@article{mccreary_calgary_2020,
	chapter = {Neurology},
	title = {Calgary {Normative} {Study}: design of a prospective longitudinal study to characterise potential quantitative {MR} biomarkers of neurodegeneration over the adult lifespan},
	volume = {10},
	copyright = {© Author(s) (or their employer(s)) 2020. Re-use permitted under CC BY-NC. No commercial re-use. See rights and permissions. Published by BMJ.. http://creativecommons.org/licenses/by-nc/4.0/This is an open access article distributed in accordance with the Creative Commons Attribution Non Commercial (CC BY-NC 4.0) license, which permits others to distribute, remix, adapt, build upon this work non-commercially, and license their derivative works on different terms, provided the original work is properly cited, appropriate credit is given, any changes made indicated, and the use is non-commercial. See: http://creativecommons.org/licenses/by-nc/4.0/.},
	issn = {2044-6055, 2044-6055},
	shorttitle = {Calgary {Normative} {Study}},
	url = {https://bmjopen.bmj.com/content/10/8/e038120},
	doi = {10.1136/bmjopen-2020-038120},
	language = {en},
	number = {8},
	urldate = {2024-12-12},
	journal = {BMJ Open},
	publisher = {British Medical Journal Publishing Group},
	author = {McCreary, Cheryl R. and Salluzzi, Marina and Andersen, Linda B. and Gobbi, David and Lauzon, Louis and Saad, Feryal and Smith, Eric E. and Frayne, Richard},
	month = aug,
	year = {2020},
	pages = {e038120},
}

@article{guarnera_mslesseg_2025,
	title = {{MSLesSeg}: baseline and benchmarking of a new {Multiple} {Sclerosis} {Lesion} {Segmentation} dataset},
	volume = {12},
	issn = {2052-4463},
	url = {https://doi.org/10.1038/s41597-025-05250-y},
	doi = {10.1038/s41597-025-05250-y},
	number = {1},
	journal = {Scientific Data},
	author = {Guarnera, Francesco and Rondinella, Alessia and Crispino, Elena and Russo, Giulia and Di Lorenzo, Clara and Maimone, Davide and Pappalardo, Francesco and Battiato, Sebastiano},
	month = may,
	year = {2025},
	pages = {920},
}

@misc{noauthor_ixi_nodate,
  title = {{IXI Dataset -- Brain Development} {http://brain-development.org/ixi-dataset/}},
  author = {{IXI Dataset}},
  year = {},
  url = {http://brain-development.org/ixi-dataset/},
}

@article{zuo_open_2014,
	title = {An open science resource for establishing reliability and reproducibility in functional connectomics},
	volume = {1},
	copyright = {2014 The Author(s)},
	issn = {2052-4463},
	url = {https://www.nature.com/articles/sdata201449},
	doi = {10.1038/sdata.2014.49},
	language = {en},
	number = {1},
	urldate = {2024-12-12},
	journal = {Scientific Data},
	publisher = {Nature Publishing Group},
	author = {Zuo, Xi-Nian and Anderson, Jeffrey S. and Bellec, Pierre and Birn, Rasmus M. and Biswal, Bharat B. and Blautzik, Janusch and Breitner, John C. S. and Buckner, Randy L. and Calhoun, Vince D. and Castellanos, F. Xavier and Chen, Antao and Chen, Bing and Chen, Jiangtao and Chen, Xu and Colcombe, Stanley J. and Courtney, William and Craddock, R. Cameron and Di Martino, Adriana and Dong, Hao-Ming and Fu, Xiaolan and Gong, Qiyong and Gorgolewski, Krzysztof J. and Han, Ying and He, Ye and He, Yong and Ho, Erica and Holmes, Avram and Hou, Xiao-Hui and Huckins, Jeremy and Jiang, Tianzi and Jiang, Yi and Kelley, William and Kelly, Clare and King, Margaret and LaConte, Stephen M. and Lainhart, Janet E. and Lei, Xu and Li, Hui-Jie and Li, Kaiming and Li, Kuncheng and Lin, Qixiang and Liu, Dongqiang and Liu, Jia and Liu, Xun and Liu, Yijun and Lu, Guangming and Lu, Jie and Luna, Beatriz and Luo, Jing and Lurie, Daniel and Mao, Ying and Margulies, Daniel S. and Mayer, Andrew R. and Meindl, Thomas and Meyerand, Mary E. and Nan, Weizhi and Nielsen, Jared A. and O’Connor, David and Paulsen, David and Prabhakaran, Vivek and Qi, Zhigang and Qiu, Jiang and Shao, Chunhong and Shehzad, Zarrar and Tang, Weijun and Villringer, Arno and Wang, Huiling and Wang, Kai and Wei, Dongtao and Wei, Gao-Xia and Weng, Xu-Chu and Wu, Xuehai and Xu, Ting and Yang, Ning and Yang, Zhi and Zang, Yu-Feng and Zhang, Lei and Zhang, Qinglin and Zhang, Zhe and Zhang, Zhiqiang and Zhao, Ke and Zhen, Zonglei and Zhou, Yuan and Zhu, Xing-Ting and Milham, Michael P.},
	month = dec,
	year = {2014},
	pages = {140049},
}

@article{di_martino_autism_2014,
	title = {The autism brain imaging data exchange: towards a large-scale evaluation of the intrinsic brain architecture in autism},
	volume = {19},
	copyright = {2014 Macmillan Publishers Limited},
	issn = {1476-5578},
	shorttitle = {The autism brain imaging data exchange},
	url = {https://www.nature.com/articles/mp201378},
	doi = {10.1038/mp.2013.78},
	language = {en},
	number = {6},
	urldate = {2024-12-12},
	journal = {Molecular Psychiatry},
	publisher = {Nature Publishing Group},
	author = {Di Martino, A. and Yan, C.-G. and Li, Q. and Denio, E. and Castellanos, F. X. and Alaerts, K. and Anderson, J. S. and Assaf, M. and Bookheimer, S. Y. and Dapretto, M. and Deen, B. and Delmonte, S. and Dinstein, I. and Ertl-Wagner, B. and Fair, D. A. and Gallagher, L. and Kennedy, D. P. and Keown, C. L. and Keysers, C. and Lainhart, J. E. and Lord, C. and Luna, B. and Menon, V. and Minshew, N. J. and Monk, C. S. and Mueller, S. and Müller, R.-A. and Nebel, M. B. and Nigg, J. T. and O'Hearn, K. and Pelphrey, K. A. and Peltier, S. J. and Rudie, J. D. and Sunaert, S. and Thioux, M. and Tyszka, J. M. and Uddin, L. Q. and Verhoeven, J. S. and Wenderoth, N. and Wiggins, J. L. and Mostofsky, S. H. and Milham, M. P.},
	month = jun,
	year = {2014},
	pages = {659--667},
}

@article{di_martino_enhancing_2017,
	title = {Enhancing studies of the connectome in autism using the autism brain imaging data exchange {II}},
	volume = {4},
	copyright = {2017 The Author(s)},
	issn = {2052-4463},
	url = {https://www.nature.com/articles/sdata201710},
	doi = {10.1038/sdata.2017.10},
	language = {en},
	number = {1},
	urldate = {2024-12-12},
	journal = {Scientific Data},
	publisher = {Nature Publishing Group},
	author = {Di Martino, A. and O’Connor, David and Chen, Bosi and Alaerts, Kaat and Anderson, Jeffrey S. and Assaf, Michal and Balsters, Joshua H. and Baxter, Leslie and Beggiato, Anita and Bernaerts, Sylvie and Blanken, Laura M. E. and Bookheimer, Susan Y. and Braden, B. Blair and Byrge, Lisa and Castellanos, F. Xavier and Dapretto, Mirella and Delorme, Richard and Fair, Damien A. and Fishman, Inna and Fitzgerald, Jacqueline and Gallagher, Louise and Keehn, R. Joanne Jao and Kennedy, Daniel P. and Lainhart, Janet E. and Luna, Beatriz and Mostofsky, Stewart H. and Müller, Ralph-Axel and Nebel, Mary Beth and Nigg, Joel T. and O’Hearn, Kirsten and Solomon, Marjorie and Toro, Roberto and Vaidya, Chandan J. and Wenderoth, Nicole and White, Tonya and Craddock, R. Cameron and Lord, Catherine and Leventhal, Bennett and Milham, Michael P.},
	month = mar,
	year = {2017},
	pages = {170010},
}

@article{elharrouss_image_2020,
	title = {Image {Inpainting}: {A} {Review}},
	volume = {51},
	issn = {1573-773X},
	shorttitle = {Image {Inpainting}},
	url = {https://doi.org/10.1007/s11063-019-10163-0},
	doi = {10.1007/s11063-019-10163-0},
	language = {en},
	number = {2},
	urldate = {2024-12-12},
	journal = {Neural Processing Letters},
	author = {Elharrouss, Omar and Almaadeed, Noor and Al-Maadeed, Somaya and Akbari, Younes},
	month = apr,
	year = {2020},
	pages = {2007--2028},
}

@inproceedings{tang_self-supervised_2022,
	address = {New Orleans, LA, USA},
	title = {Self-{Supervised} {Pre}-{Training} of {Swin} {Transformers} for {3D} {Medical} {Image} {Analysis}},
	copyright = {https://doi.org/10.15223/policy-029},
	isbn = {978-1-6654-6946-3},
	url = {https://ieeexplore.ieee.org/document/9879123/},
	doi = {10.1109/CVPR52688.2022.02007},
	language = {en},
	urldate = {2025-01-05},
	booktitle = {2022 {IEEE}/{CVF} {Conference} on {Computer} {Vision} and {Pattern} {Recognition} ({CVPR})},
	publisher = {IEEE},
	author = {Tang, Yucheng and Yang, Dong and Li, Wenqi and Roth, Holger R. and Landman, Bennett and Xu, Daguang and Nath, Vishwesh and Hatamizadeh, Ali},
	month = jun,
	year = {2022},
	pages = {20698--20708},
}

@inproceedings{hatamizadeh_unetr_2022,
	title = {{UNETR}: {Transformers} for {3D} {Medical} {Image} {Segmentation}},
	shorttitle = {{UNETR}},
	url = {https://openaccess.thecvf.com/content/WACV2022/html/Hatamizadeh_UNETR_Transformers_for_3D_Medical_Image_Segmentation_WACV_2022_paper.html},
	language = {en},
	urldate = {2025-01-05},
	author = {Hatamizadeh, Ali and Tang, Yucheng and Nath, Vishwesh and Yang, Dong and Myronenko, Andriy and Landman, Bennett and Roth, Holger R. and Xu, Daguang},
	year = {2022},
	pages = {574--584},
}

@inproceedings{ronneberger_u-net_2015,
	address = {Cham},
	title = {U-{Net}: {Convolutional} {Networks} for {Biomedical} {Image} {Segmentation}},
	isbn = {978-3-319-24574-4},
	shorttitle = {U-{Net}},
	doi = {10.1007/978-3-319-24574-4_28},
	language = {en},
	booktitle = {Medical {Image} {Computing} and {Computer}-{Assisted} {Intervention} – {MICCAI} 2015},
	publisher = {Springer International Publishing},
	author = {Ronneberger, Olaf and Fischer, Philipp and Brox, Thomas},
	editor = {Navab, Nassir and Hornegger, Joachim and Wells, William M. and Frangi, Alejandro F.},
	year = {2015},
	pages = {234--241},
}

@misc{zheng_self-supervised_2024,
	title = {Self-{Supervised} {Pretext} {Tasks} for {Alzheimer}'s {Disease} {Classification} using {3D} {Convolutional} {Neural} {Networks} on {Large}-{Scale} {Synthetic} {Neuroimaging} {Dataset}},
	url = {http://arxiv.org/abs/2406.14210},
	doi = {10.48550/arXiv.2406.14210},
	urldate = {2025-03-13},
	publisher = {arXiv},
	author = {Zheng, Chen},
	month = jun,
	year = {2024},
	note = {arXiv:2406.14210 [eess]},
}

@incollection{cho_domain_2025,
	address = {Singapore},
	title = {Domain {Aware} {Multi}-task {Pretraining} of {3D} {Swin} {Transformer} for {T1}-{Weighted} {Brain} {MRI}},
	volume = {15473},
	isbn = {978-981-96-0900-0 978-981-96-0901-7},
	url = {https://link.springer.com/10.1007/978-981-96-0901-7_8},
	doi = {10.1007/978-981-96-0901-7_8},
	language = {en},
	urldate = {2025-03-13},
	booktitle = {Computer {Vision} – {ACCV} 2024},
	publisher = {Springer Nature Singapore},
	author = {Kim, Jonghun and Kim, Mansu and Park, Hyunjin},
	editor = {Cho, Minsu and Laptev, Ivan and Tran, Du and Yao, Angela and Zha, Hongbin},
	year = {2025},
	note = {Series Title: Lecture Notes in Computer Science},
	pages = {121--141},
}

@inproceedings{stoean_using_2022,
	title = {On {Using} {Perceptual} {Loss} within the {U}-{Net} {Architecture} for the {Semantic} {Inpainting} of {Textile} {Artefacts} with {Traditional} {Motifs}},
	issn = {2470-881X},
	url = {https://ieeexplore.ieee.org/abstract/document/10131036},
	doi = {10.1109/SYNASC57785.2022.00051},
	urldate = {2025-03-24},
	booktitle = {2022 24th {International} {Symposium} on {Symbolic} and {Numeric} {Algorithms} for {Scientific} {Computing} ({SYNASC})},
	author = {Stoean, Catalin and Bacanin, Nebojsa and Stoean, Ruxandra and Ionescu, Leonard and Alecsa, Cristian and Hotoleanu, Mircea and Atencia, Miguel and Joya, Gonzalo},
	month = sep,
	year = {2022},
	pages = {276--283},
}

@inproceedings{zhang_unreasonable_2018,
	title = {The unreasonable effectiveness of deep features as a perceptual metric},
	url = {http://openaccess.thecvf.com/content_cvpr_2018/html/Zhang_The_Unreasonable_Effectiveness_CVPR_2018_paper.html},
	urldate = {2025-03-24},
	booktitle = {Proceedings of the {IEEE} conference on computer vision and pattern recognition},
	author = {Zhang, Richard and Isola, Phillip and Efros, Alexei A. and Shechtman, Eli and Wang, Oliver},
	year = {2018},
	pages = {586--595},
}

@article{jack_alzheimers_2008,
	title = {The {Alzheimer}'s disease neuroimaging initiative ({ADNI}): {MRI} methods},
	volume = {27},
	copyright = {http://onlinelibrary.wiley.com/termsAndConditions\#vor},
	issn = {1053-1807, 1522-2586},
	shorttitle = {The {Alzheimer}'s disease neuroimaging initiative ({ADNI})},
	url = {https://onlinelibrary.wiley.com/doi/10.1002/jmri.21049},
	doi = {10.1002/jmri.21049},
	language = {en},
	number = {4},
	urldate = {2025-03-24},
	journal = {Journal of Magnetic Resonance Imaging},
	author = {Jack, Clifford R. and Bernstein, Matt A. and Fox, Nick C. and Thompson, Paul and Alexander, Gene and Harvey, Danielle and Borowski, Bret and Britson, Paula J. and L. Whitwell, Jennifer and Ward, Chadwick and Dale, Anders M. and Felmlee, Joel P. and Gunter, Jeffrey L. and Hill, Derek L.G. and Killiany, Ron and Schuff, Norbert and Fox‐Bosetti, Sabrina and Lin, Chen and Studholme, Colin and DeCarli, Charles S. and {Gunnar Krueger} and Ward, Heidi A. and Metzger, Gregory J. and Scott, Katherine T. and Mallozzi, Richard and Blezek, Daniel and Levy, Joshua and Debbins, Josef P. and Fleisher, Adam S. and Albert, Marilyn and Green, Robert and Bartzokis, George and Glover, Gary and Mugler, John and Weiner, Michael W.},
	month = apr,
	year = {2008},
	pages = {685--691},
}

@article{shafto_cambridge_2014,
	title = {The {Cambridge} {Centre} for {Ageing} and {Neuroscience} ({Cam}-{CAN}) study protocol: a cross-sectional, lifespan, multidisciplinary examination of healthy cognitive ageing},
	volume = {14},
	copyright = {2014 Shafto et al.; licensee BioMed Central Ltd.},
	issn = {1471-2377},
	shorttitle = {The {Cambridge} {Centre} for {Ageing} and {Neuroscience} ({Cam}-{CAN}) study protocol},
	url = {https://link.springer.com/article/10.1186/s12883-014-0204-1},
	doi = {10.1186/s12883-014-0204-1},
	language = {en},
	number = {1},
	urldate = {2025-03-24},
	journal = {BMC Neurology},
	publisher = {BioMed Central},
	author = {Shafto, Meredith A. and Tyler, Lorraine K. and Dixon, Marie and Taylor, Jason R. and Rowe, James B. and Cusack, Rhodri and Calder, Andrew J. and Marslen-Wilson, William D. and Duncan, John and Dalgleish, Tim and Henson, Richard N. and Brayne, Carol and Matthews, Fiona E.},
	month = dec,
	year = {2014},
	note = {Number: 1},
	pages = {1--25},
}

@article{he_foundation_2025,
	title = {Foundation {Model} for {Advancing} {Healthcare}: {Challenges}, {Opportunities} and {Future} {Directions}},
	volume = {18},
	issn = {1941-1189},
	shorttitle = {Foundation {Model} for {Advancing} {Healthcare}},
	url = {https://ieeexplore.ieee.org/document/10750441/},
	doi = {10.1109/RBME.2024.3496744},
	urldate = {2025-04-30},
	journal = {IEEE Reviews in Biomedical Engineering},
	author = {He, Yuting and Huang, Fuxiang and Jiang, Xinrui and Nie, Yuxiang and Wang, Minghao and Wang, Jiguang and Chen, Hao},
	year = {2025},
	pages = {172--191},
}

@inproceedings{pathak_context_2016,
	address = {Las Vegas, NV, USA},
	title = {Context {Encoders}: {Feature} {Learning} by {Inpainting}},
	isbn = {978-1-4673-8851-1},
	shorttitle = {Context {Encoders}},
	url = {http://ieeexplore.ieee.org/document/7780647/},
	doi = {10.1109/CVPR.2016.278},
	language = {en},
	urldate = {2025-05-03},
	booktitle = {2016 {IEEE} {Conference} on {Computer} {Vision} and {Pattern} {Recognition} ({CVPR})},
	publisher = {IEEE},
	author = {Pathak, Deepak and Krahenbuhl, Philipp and Donahue, Jeff and Darrell, Trevor and Efros, Alexei A.},
	month = jun,
	year = {2016},
	pages = {2536--2544},
}

@inproceedings{nasser_investigating_2026,
	address = {Cham},
	title = {Investigating {Voxel}-{Level} {Brain} {Age} {Prediction} as a {Pretext} {Task} for {Brain} {MRI} {Segmentation}},
	isbn = {978-3-032-05325-1},
	doi = {10.1007/978-3-032-05325-1_28},
	language = {en},
	booktitle = {Medical {Image} {Computing} and {Computer} {Assisted} {Intervention} – {MICCAI} 2025},
	publisher = {Springer Nature Switzerland},
	author = {Nasser, Tasneem and Souza, Roberto and El-Sheimy, Naser},
	editor = {Gee, James C. and Alexander, Daniel C. and Hong, Jaesung and Iglesias, Juan Eugenio and Sudre, Carole H. and Venkataraman, Archana and Golland, Polina and Kim, Jong Hyo and Park, Jinah},
	year = {2026},
	pages = {289--299},
}


\end{document}